\let\LaTeXcline\cline\documentclass[5p,times,twocolumn]{elsarticle}\let\cline\LaTeXcline

\usepackage[utf8]{inputenc} %
\usepackage[T1]{fontenc}    %

\usepackage{hyperref}       %

\usepackage{url}            %
\usepackage{booktabs}       %
\usepackage{amsfonts}       %
\usepackage{nicefrac}       %
\usepackage{microtype}      %
\usepackage[dvipsnames]{xcolor}         %
\usepackage{amsmath}
\usepackage{amssymb}
\usepackage{array}
\usepackage{amsthm}
\usepackage{svg}
\usepackage{bbm}
\usepackage{bm}
\usepackage{algorithm}
\usepackage{algpseudocode}
\usepackage{cleveref}
\usepackage[normalem]{ulem}
\usepackage{comment}
\usepackage{multirow}

\usepackage{nkj}
\usepackage{enumitem}

\usepackage{caption}
\usepackage{subcaption}

\newcommand{\numneu}[1]{{N^{(#1)}}}

\newcommand{\ol}{L}

\newcommand{\il}{0}

\newcommand{\subg}[1]{ {\mcS}^{(#1)}}

\newcommand{\eventtime}[2][1em]{%
  $\overset{#2}{\raisebox{0.5ex}{\rule{#1}{0.5pt}}}$%
}

\ifdefined\QED
\else
\newcommand{\QED}{\hfill \ensuremath{\Box}}
\fi

\makeatletter
\renewcommand\paragraph{\@startsection{paragraph}{4}{\z@}%
  {3.25ex \@plus1ex \@minus.2ex}%
  {-1em}%
  {\normalfont\normalsize\bfseries\unskip}}
\makeatother

\usepackage{accents}

\usepackage{colortbl}

\begin{document}

\begin{frontmatter}

  \title{Explaining Temporal Graph Neural Networks via Feature-induced Information Flow}

 \author[1,2]{Ping Xiong}
\author[7,8,9]{Thomas Schnake}
\author[1,2,4,5]{Klaus-Robert M\"uller}
\author[1,2,3]{Shinichi Nakajima\corref{cor}}
\ead{nakajima@tu-berlin.de}

\cortext[cor]{Corresponding author
}

\address[1]{Berlin Institute for the Foundations of Learning and Data -- BIFOLD, 10623 Berlin, Germany}
\address[2]{Machine Learning Group, Technical University of Berlin, 
Berlin, Germany}
\address[7]{Department of Chemistry, Chemical Physics Theory Group, University of Toronto, Toronto, 
Canada}
\address[8]{Vector Institute for Artificial Intelligence, Toronto, 
Canada}
\address[9]{Acceleration Consortium, University of Toronto, Toronto, 
Canada}
\address[4]{Department of Artificial Intelligence, Korea University, Seoul,
Korea}
\address[5]{Max Planck Institute for Informatics, 
Saarbr\"ucken, Germany}
\address[3]{RIKEN AIP, 
Tokyo, Japan}

\begin{abstract}
Event-based Temporal Graph Neural Networks (ETGNNs) have demonstrated strong performance across a wide range of applications, including social network analysis, epidemic tracing, recommender systems, and political event forecasting.  However, their increasing complexity poses significant challenges for explainability.
Existing explanation methods focus only on a subset of the information flow within ETGNNs, typically tracing contributions from the event-related embeddings to the output.  Consequently, they overlook the important pathways through event-induced variables, which mediate interactions between nodes and thereby play a central role in capturing long-range temporal dependencies.
To overcome this limitation, we propose a novel attribution method that analyzes the \emph{entire} information flow through all event-associated variables.
Our method is built upon the recent  Normalized Relevance Measure (NRM) framework, which enables explicit quantification of information flow originating from event embeddings as well as information flow passing through event-induced variables.  It also ensures comparability of latent variables across layers, and supports higher-order  analysis of interactions between events. 
To handle the architectural complexity of ETGNNs, we extend the NRM framework with a modular decomposition procedure that facilitates the systematic construction of relevance structure for complex neural architectures.
We evaluate our approach on two synthetic datasets for epidemic tracing and social dynamics, as well as a real-world dataset of political event networks.
Our qualitative and quantitative experiments show that our method consistently outperforms existing explanation approaches while producing more human-interpretable explanations.
\end{abstract}

\begin{keyword}
explainable artificial intelligence, layer-wise relevance propagation, normalized signed measure, temporal graph neural networks
\end{keyword}
\end{frontmatter}

\begin{figure*}[t]
    \begin{center}
        \includegraphics[width=\textwidth]{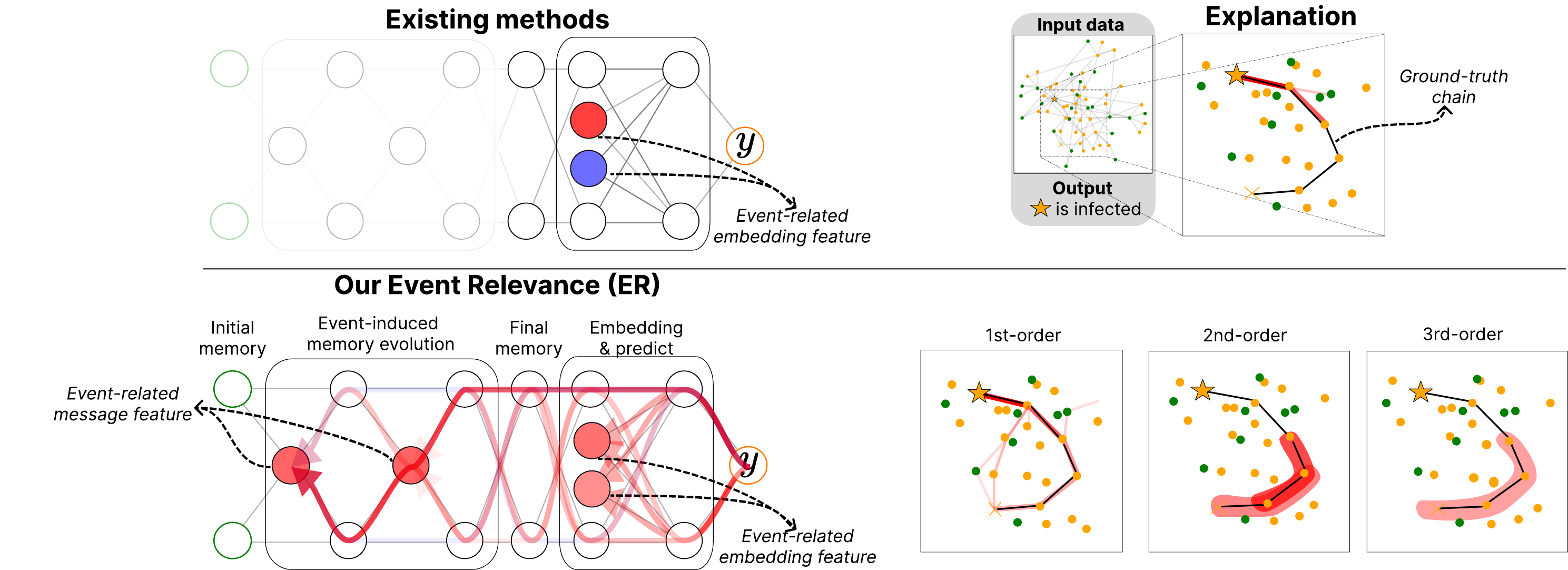}
    \end{center}
    \caption{
    Overview of the proposed Event Relevance (ER). 
    Top row: Existing explanation methods primarily analyze the final embedding and prediction stages, overlooking the latent memory evolution process induced by time-stamped events.  Consequently, they often fail to identify the complete event chain, as illustrated on the right.
    Bottom row: Our ER accounts for the complete information flow throughout the entire ETGNN architecture, thereby identifying the event chain that contribute to the prediction and providing more faithful and fine-grained explanations. 
    Furthermore, it naturally extends to higher-order joint event relevance, enabling the discovery of interactions among multiple events.
    }
    \label{fig:fig1_expl_concept}
\end{figure*}

\section{Introduction}
\label{sec:intro}

Event-based Temporal Graph Neural Networks (ETGNNs) are powerful models for learning from dynamic relational data represented as temporal graphs---where edges are induced by events with time-stamps---and have been applied across a wide range of domains, including recommender systems \citep{DBLP:conf/wsdm/GaoW0022}, social networks \citep{DBLP:conf/kdd/DengRN19}, epidemic modeling \citep{cencetti_digital_2021}, and traffic networks \citep{DBLP:journals/tits/ZhaoSZLWLDL20}. 
However, 
ETGNNs pose significant challenges in explainability, as their complex architectures make it difficult to determine what information should be attributed and how attribution should be performed.
This lack of interpretability is particularly problematic in high-stakes applications, where understanding the rationale behind model predictions is essential.

Several explainable artificial intelligence (XAI) methods, such as Temporal GNN Explainer (TGNNExplainer) \citep{DBLP:conf/iclr/XiaLS0D0023} and Temporal Motifs Explainer (TempME) \citep{DBLP:conf/nips/ChenY23}, 
have been proposed, and have successfully identified relevant events for predictions in social networks and political event networks.  However, they primarily explain the later stages of the network, while overlooking the upstream computational process that governs the temporal evolution of the graph. As a result, their explanatory power is limited, since long-range temporal dependencies play a central role in ETGNN predictions.

To address this limitation, we propose a novel explanation method that considers information flow throughout the entire ETGNN.  Specifically, our definition of event relevance (ER) accounts for the entire information flow associated with an event, including flow passing through intermediate features, called \emph{messages}, that are induced by events and mediate interactions between nodes. Analyzing these messages is crucial for understanding how ETGNNs capture long-range temporal dependencies.

We build our method upon the \emph{Normalized Relevance Measure} (NRM) framework \cite{NRMpaper}, which provides a principled way to define the relevance of arbitrary sets of neurons and derive the corresponding Layer-wise Relevance Propagation (LRP) algorithms \cite{bach2015pixel}.
NRM generalizes GNN-LRP \citep{schnake2020higher}, a higher-order XAI method originally developed for GNN, together with its efficient implementation \citep{DBLP:conf/icml/XiongSMMN22}.
This framework enables explicit quantification of the complete event-induced information flow while guaranteeing the comparability of relevance scores across layers.  Furthermore, it supports the joint relevance analysis, which quantifies the contribution of higher-order interactions among events.  This capability is particularly important in applications such as infection chain analysis, where sequences of events may jointly contribute to the predicted outcome.

To handle the complex architecture of ETGNNs, we propose a hierarchical procedure for defining relevance via \emph{modularization}, which significantly simplifies the application of the NRM framework to complex networks. 
Building upon this extension of NRM framework, 
our ER captures information flow more faithfully throughout the entire model, including the latent memory evolution process, which is overlooked by existing methods. Furthermore, our joint event relevance identifies important higher-order interactions, providing a more fine-grained explanation of information flow. \Cref{fig:fig1_expl_concept} provides a conceptual illustration of our approach.

In our experiments, we constructed synthetic datasets with ground-truth explanation that simulate disease infections and social networks, enabling human-verifiable evaluation of XAI methods.  Our qualitative evaluation on these synthetic datasets, as well as on a real-world dataset of political events,
demonstrate that our proposed ER yields human-interpretable explanations.
Our quantitative experiments further validate
the superior performance of our approach compared to existing XAI methods.

\medskip

The main contributions of this paper are summarized as follows:
\begin{itemize}

\item 
We propose a novel ER that attributes predictions to each event based on information flow throughout the entire~\mbox{ETGNN}.

\item 
We extend the NRM framework with a modular decomposition procedure 
for hierarchically defining relevance, which significantly simplifies its application to complex networks.

\item 
We demonstrate the superior performance of our approach, both qualitatively and quantitatively, compared to existing baseline methods across different prediction tasks and~datasets.

\end{itemize}

The remainder of the paper is organized as follows. After reviewing related work in \Cref{sec:RelatedWork}, we 
present our modularization-based extension of the NRM framework in \Cref{sec:Module_NRM}. We then propose our ER definitions for ETGNNs in \Cref{sec:ExplainETGNN}, followed by experimental evaluations in \Cref{sec:Experiment}. \Cref{sec:Conclusion} concludes.

\section{Related Work}
\label{sec:RelatedWork}

In this section, we review related work on LRP, NRM, ETGNNs, and existing XAI methods for ETGNNs. 
Comprehensive surveys of general XAI methods and their evaluation are available in recent literature \cite{DBLP:journals/inffus/ArrietaRSBTBGGM20, 
DBLP:journals/inffus/AliAEMACGSRH23}.

\subsection{Layer-wise Relevance Propagation (LRP)
}

Layer-wise Relevance Propagation (LRP) \citep{bach2015pixel} is a post-hoc, propagation-based XAI method for neural networks, 
which decomposes model predictions into relevance scores of input features via a backward propagation from output to input. As it requires only a single backward pass, LRP is computationally efficient.
LRP is mathematically justified as a first-order Deep Taylor Decomposition (DTD) \citep{DBLP:journals/pr/MontavonLBSM17, DBLP:journals/pieee/SamekMLAM21}, approximating the forward process with layer-wise linearizations.
Bach et al.~\cite{bach2015pixel} introduced LRP with propagation rules such as $\varepsilon$-rule, which were
later extended by the $\gamma$-rule \citep{DBLP:series/lncs/MontavonBLSM19} and further adapted for recurrent networks \citep{DBLP:series/lncs/ArrasAWMGMHS19}. LRP has been widely applied in various fields, including computer vision \citep{AttnLRP, kauffmann2025explainable, lapuschkin2019unmasking}, natural language processing \citep{arras2017textlrp, DBLP:conf/icml/AliSEMMW22}, and quantum chemistry \citep{schnake2020higher, esders2025long_range_xai}.
It has also been used for model debugging and identifying spurious decision strategies \cite{DBLP:journals/inffus/AndersWNSML22, DBLP:journals/inffus/WeberLBS23, DBLP:journals/inffus/BenderDHMMM26}.

LRP has also been applied to analyze information flow across layers. Schnake et al.~\cite{schnake2020higher} introduced the notion of a \emph{walk}, a sequence of neurons from the input layer to the output layer, and defined its relevance based on LRP propagation rules.  Building on this concept, the authors proposed GNN-LRP, which explains GNN predictions in terms of walk relevances.  Specifically, GNN-LRP attributes to a  given subgraph the total relevance of the walks passing through neuron sets associated with the nodes in the subgraph.  Here the associated neuron set consists of neurons representing the node embeddings across layers.
Our approach follows a similar spirit for event-level explanations for ETGNNs, where each event is attributed the total relevance of the walks passing through the neuron set associated with that event.

\subsection{Normalized Relevance Measure (NRM)}
Xiong et al.~\cite{NRMpaper} proposed the Normalized Relevance Measure (NRM), a general framework for defining relevance of arbitrary neuron sets and deriving the corresponding LRP algorithms.  This framework is motivated by the similarity between LRP computations and message passing probability computation in Markov chains \citep{DBLP:conf/icml/XiongSMMN22}, and allows us to directly specify the relevance quantity to be computed.  The relevance specification is carried out via marginalization and conditional operations, in analogy to the probability theory.
Furthermore, NRM guarantees comparability of neurons across layers, and enables joint relevance analysis to capture higher-order interactions.
Our method is built on NRM to capture entire event-induced information flow, and to provide higher-order explanations.

\subsection{Temporal Graph Neural Networks (TGNN)}
\label{sec:BG.TGNNExp}

Graphs are an important data representation in many application domains, including social networks \citep{chen2018fastgcn, hamilton2017inductive, DBLP:conf/iclr/KipfW17}, molecular dynamics \citep{schutt2018schnet}, and natural language processing \citep{yuan2020explainability}.
Consequently, a wide range of graph neural networks (GNNs) have been proposed
\citep{DBLP:journals/nn/JuFGLLQQSSXYYZWLZ24}.
GNNs have also been investigated as a means of explainable and causability-oriented analysis \cite{DBLP:journals/inffus/HolzingerMSP21}.
Analyzing temporal graphs, i.e., temporally evolving graphs, has recently drawn attentions in the research fields of recommender systems \citep{DBLP:conf/wsdm/GaoW0022}, social networks \citep{DBLP:conf/kdd/DengRN19}, epidemic modeling \citep{cencetti_digital_2021}, and traffic networks \citep{DBLP:journals/tits/ZhaoSZLWLDL20}, where capturing the dynamics of graph evolutions plays an important role.
Although GNNs have been adapted for handling temporal graphs \citep{hamilton2017inductive,DBLP:conf/cvpr/MontiBMRSB17}, 
more specialized architectures, called 
Temporal Graph Neural Networks (TGNNs) \citep{DBLP:journals/corr/abs-2006-10637,DBLP:journals/tits/ZhaoSZLWLDL20},
have shown to improve the performance.
In this paper, 
among many variants of TGNNs, we focus on Event-based TGNNs (ETGNN)\footnote{
    Although ETGNN is named Temporal Graph Network (TGN)
    in the original paper \citep{DBLP:journals/corr/abs-2006-10637},
    we call it ETGNN throughout this paper to avoid confusion with other TGNNs.
    }
\citep{DBLP:journals/corr/abs-2006-10637,DBLP:conf/sigir/0001GRTY20}---%
one of the most general architectures including 
Jodie \citep{DBLP:conf/kdd/KumarZL19}, TGAT \citep{DBLP:conf/iclr/XuRKKA20}, and DyRep \citep{DBLP:conf/iclr/TrivediFBZ19}.

In ETGNNs, a temporal graph is defined as an initial graph and a series of \emph{events}, where
each event 
induces an interaction between two nodes at a specific time, dynamically adding/removing the corresponding link %
at the time.
An \mbox{ETGNN} generally consists of three modules, Event Processing (EP), Embedding (Emb), and Decoding (Dec) modules.
The EP module is normally a recurrent neural network (RNN) that updates the time-dependent node-embeddings, called \emph{memory}, according to the associated events.  
Using the memories and the temporal graph at the inference time, the Emb module finalizes the node representations, and the Dec module makes predictions for downstream tasks, e.g., node regression, link prediction, and graph classification. 

\subsection{XAI for ETGNNs} 

Several methods have been proposed for explaining ETGNNs.
Temporal GNN Explainer (TGNNExplainer) \citep{DBLP:conf/iclr/XiaLS0D0023} applies Monte-Carlo Tree Search (MCTS) \citep{DBLP:conf/ecml/KocsisS06} to find the subgraph---as a set of event edges---that dominate the network decisions. The search starts with a subgraph of a given number of most recent neighboring events, and the MCTS removes unimportant events one-by-one from the subgraph, until the target size of subgraph is reached. 
The search process is accelerated by an MLP that predicts the edge importance score.
Temporal Motifs Explainer (TempME) \citep{DBLP:conf/nips/ChenY23} samples \emph{temporal graph motifs} and train an MLP to predict the motifs' importance scores, which are used to build an important subgraph. The method uses information bottleneck theory \citep{
DBLP:journals/corr/physics-0004057} to optimize the MLP such that the important subgraph generates similar prediction as the original input graph while being small.

These existing XAI methods offer valuable insights into the model's prediction mechanisms.  However, they primarily focus on the Emb and Dec modules, overlooking the  information flow within the EP module.  This limitation is crucial, as long-time dependencies are typically captured within the EP module.
To address this limitation, we propose an alternative LRP-based XAI method that accounts for the information flow in all modules. %
We show 
in \Cref{sec:Experiment}
that our proposed methods provides more faithful explanations than TGNNExplainer \citep{DBLP:conf/iclr/XiaLS0D0023} and basic baseline methods.%
\footnote{
We were unable to include TempME \citep{DBLP:conf/nips/ChenY23} in our experimental comparison
due to known reproducibility issues with its official codebase (\url{https://github.com/Graph-and-Geometric-Learning/TempME/issues/3}).  We expect TempME to exhibit similar weaknesses to TGNNExplainer, as both methods ignore the information flow within the EP module.
}

\section{Modular Decomposition for NRM}
\label{sec:Module_NRM}

Our method is built upon the NRM framework \citep{NRMpaper}, which supports direct specification of the relevance quantity, comparability across layers, and higher-order analysis.  However, applying it to complex architectures requires defining the relevance structure over a large network, which can be challenging. To facilitate this process, we introduce a modular decomposition procedure that enables hierarchical relevance definition.

\subsection{Overview of NRM}

Let us consider an $L$-layer feed forward neural network (FFNN) to be a function $\bff: \mathbb{R}^{\numneu{\il}} \mapsto \mathbb{R}^{\numneu{\ol}}$ that maps from an $N^{(0)}$-dimensional input vector to a $N^{(L)}$-dimensional output vector. For each intermediate layer 
$l = 1, \dots, L-1 $, there is a latent vector that consists of $N^{(l)}$ neurons. Let $\mathbb{N}^{(l)} \equiv \{ 1, \dots, N^{(l)} \}$ denote the set of neuron indices in layer $l$. 

In the Normalized Relevance Measure (NRM) framework \citep{NRMpaper}, the relevance of a set of walks is defined as a normalized signed measure, satisfying $
    R^{\mathrm{Walk}}( \emptyset)  = 0$, $
    R^{\mathrm{Walk}}( \mathbb{W}) = 1,  
R^{\mathrm{Walk}}(\bigcup_{j} \mcW_j)
=\textstyle
\sum_{j} R^{\mathrm{Walk}}(\mcW_j)$  for any pairwise disjoint sets $\mcW_j \in \mathcal{P}(\mathbb{W})$.
Here, $\mathbb{W} \equiv   \otimes_{l = 0}^L \, \mathbb{N}^{(l)}$ is the set of walks, and $\mcP(\mathbb{W})$ is its power set.
Then, we identify the relevance of the set of neurons with the relevance of all walks that pass through the specified sets, 
i.e.,
\begin{align}
R(\mathcal{S}^{(\mcL)} )
 &=
R^{\mathrm{Walk}}(\otimes_{l = 0}^L \, \widetilde{\mcS}^{(l)}),
\notag \\
 \mbox{ where } & \quad
\widetilde{\mcS}^{(l)}
 = \textstyle \begin{cases}
  \mcS^{(l)} & \mbox{ for } l \in \mcL, \\
  \mathbb{N}^{(l)}  & \mbox{ for } l \notin \mcL.
\end{cases}
\notag
\end{align}
Here, $\mathbb{L} \equiv \{ 0, \dots, L \}$,
$\mcL =\{l_1, \ldots, l_{|\mcL|}\} \subset \mathbb{L}$, and
we represent a set of neurons as a sequence of neuron sets over selected layers, i.e., $\mcS^{(\mcL)} = (\mcS^{(l)})_{l \in \mcL}$.

Under this definition, the relevance of any set of neurons is expressed as a sum of walk relevances $R(\bfn) = R(n^{(0)}, \ldots, n^{(L)})$ with $n^{(l)} \in \mathbb{N}^{(l)}$, i.e.,
\begin{align}
    R(\mcS^{(\mcL)})
& = \textstyle \sum_{ \bfn \in \mathbb{W} :\, n^{(l)} \in \widetilde{\mcS}^{(l)}  \forall l \in \mcL } R(\bfn).
\notag
\end{align}

The relevance measure $R(\cdot)$ defined this way satisfies all properties of a probability measure $P(\cdot)$ except for non-negativity.%
\footnote{Under this analogy, layers, neurons, and walks correspond to random variables, their possible values, and joint assignments, respectively.}  Accordingly, we adopt notation from probability theory, and define joint, marginal, conditional relevances, in analogy to their probabilistic counterparts, e.g.,
\begin{align}
      R(n^{(l_1)}, n^{(l_2)}) &= \textstyle \sum_{n^{(l)} \in \mathbb{N}^{(l)} \forall l \notin \{l_1, l_2\}  }    R(\bfn),
      \label{eq:Ex.JointRelevance}\\
      R(n^{(l_2)}) &= \textstyle \sum_{n^{(l_1)} \in \mathbb{N}^{(l_1)}  }    R(n^{(l_1)}, n^{(l_2)}),
            \label{eq:Ex.MarginalRelevance}\\
      R(n^{(l_1)}| n^{(l_2)}) &= \textstyle 
     \begin{cases}
        \frac{R(n^{(l_1)}, n^{(l_2)})}{R(n^{(l_2)})} & \mbox{if } R(n^{(l_2)}) \ne 0,\\
        0 & \mbox{otherwise}.
     \end{cases}
            \label{eq:Ex.ConditionalRelevance}
\end{align}
Under the backward Markov property, 
the walk relevance, which corresponds to the full joint relevance, is decomposed as
\begin{align}
R(\bfn) = \textstyle \left(\prod_{l=1}^L R(n^{(l-1)} | n^{(l)})\right) R(n^{(L)}).
\label{eq:MarkovDecomposition}
\end{align}
Specifying each factor on the right-hand side of Eq.\eqref{eq:MarkovDecomposition}, based on propagation rules and network outputs, defines the relevance structure of the entire network.

Although the NRM framework assumes 
a restricted class of FFNNs, called \emph{proper} FFNNs, 
any architecture can be virtually converted into  a \emph{proper} FFNN \citep{NRMpaper}, and thus the framework is applicable to genereal architectures.

For introducing modularization in the next subsection, we use the notion of a \emph{substructure} and its relevance:
For a set of consecutive layers, i.e., $\mcL = \{ \underline{l}, \underline{l}+1, \dots, \overline{l}-1, \overline{l} \}$, we call the set of neurons $\mcS^{(\mcL)} = \mcS^{(\underline{l}:\overline{l})} = (\mcS^{(\underline{l})}, \dots, \mcS^{(\overline{l})})$ a substructure, and its relevance 
 \begin{align}
R( \mcS^{(\underline{l}:\overline{l})})
  & =  \textstyle
 \sum_{n^{(\underline{l})} \in \subg{\underline{l}}, \ldots, n^{(\overline{l})} \in \subg{\overline{l}}}   
    R(n^{(\underline{l})}, \ldots, n^{(\overline{l})} )
  \label{eq:A.SubstructureRelevance}
  \end{align}
a substructure relevance.

\subsection{Modularization}
\label{sec:Modularization.MOdularization}
NNs with complicated architectures typically consist of separable components.  When explaining such models, it is often convenient to define relevance locally within each component.  
We define a substructure $\mcS^{(\underline{l}:\overline{l})} = (\subg{\underline{l}}, \ldots, \subg{\overline{l}})$ to be a \emph{module}
if all interactions with the remainder of the network occur exclusively through its input and output layers.
Namely, a module is connected to other components only through neurons in $ \{\subg{\underline{l}}, \subg{\overline{l}}\}$. 
Note that a module can share their input and output neurons with other modules.
    
We also define a \emph{local relevance} and a \emph{local conditional relevance} with respect to a substructure $\mcS^{(\underline{l}:\overline{l})}$ as the renormalized relevance within the substructure, i.e.,
 \begin{align}
 R^{ \subseteq \mcS^{(\underline{l}:\overline{l})}}
(\bfn^{(\underline{l}:\overline{l})} )
&= \textstyle
\frac{
\mathbbm{1}(
\bfn^{(\underline{l}:\overline{l})} \in 
\mcS^{(\underline{l}:\overline{l})})
}
{
R(
\mcS^{(\underline{l}:\overline{l})})
}
R
(\bfn^{(\underline{l}:\overline{l})}),
  \label{eq:LocalRelevance}
  \end{align}
   where 
   $\bfn^{(\underline{l}:\overline{l})} = (n^{(\underline{l})}, \ldots, n^{(\overline{l})})$ denotes a partial walk, and 
$\mathbbm{1}( \cdot)$ is the indicator function equal to one if the event is true and zero otherwise. 
Local joint, marginal, conditional relevances are similarly defined as Eqs.\eqref{eq:Ex.JointRelevance}--\eqref{eq:Ex.ConditionalRelevance}, respectively, by replacing the full joint relevance $R(\bfn)$ with its locally normalized variant \eqref{eq:LocalRelevance}.

\begin{figure}[!t]
    \centering
   \includegraphics[width=0.8\linewidth]{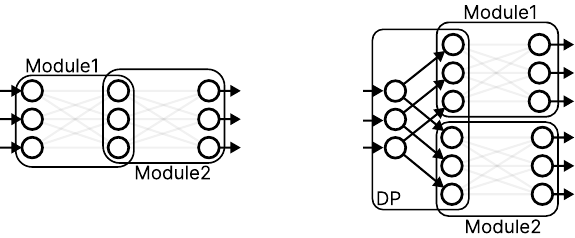}
   \caption{ A series connection (left) and a parallel connection (right) of modules. Parallel connections often require the (virtual) insertion of a DuPlication (DP) layer so that both parallel modules can receive the forward information from the same input neurons, while ensuring separability.
   }
    \label{fig:A.Modules}
\end{figure}

The multiplication and addition laws of relevance measure \citep{NRMpaper} naturally give rise to the following modular decompositions.

\paragraph{Series Connections of Modules}
Assume that the layers $\underline{l}:\overline{l}$
can be divided into two modules connected in series (see \Cref{fig:A.Modules} left).  If the first module consists of the layers $\underline{l}:l$ and the second module consists of the layers $l:\overline{l}$, the relevance can be decomposed as
\begin{align}
R(\bfn^{(\underline{l}:\overline{l}-1)}| n^{(\overline{l})}) 
&=
R(\bfn^{(\underline{l}:l-1)}| n^{(l)}) R(\bfn^{(l:\overline{l}-1)}| n^{(\overline{l})} ).
\label{eq:SeriesDecompositionConditional} 
\end{align}

\paragraph{Parallel Connections of Modules}

Assume that the substructures $\mcS^{(\underline{l}:\overline{l})} = (\subg{\underline{l}}, \ldots, \subg{\overline{l}})$ 
and
$\overline{\mcS}^{(\underline{l}:\overline{l})} = (\overline{\mcS}^{(\underline{l})}, \ldots,
\overline{\mcS}^{(\overline{l})})$
are \emph{non-overlapping} modules in parallel that are \emph{complement} of each other
 (see \Cref{fig:A.Modules} right).
Then, 
it holds that 
\begin{align}
R(\bfn^{(\underline{l}:\overline{l}-1)} | n^{(\overline{l})}) 
&=
R^{ \subseteq \mcS^{(\underline{l}:\overline{l})}}
(\bfn^{(\underline{l}:\overline{l}-1)}  | n^{(\overline{l})} )
\notag\\
& \qquad \qquad
+
R^{ \subseteq \overline{\mcS}^{(\underline{l}:\overline{l})}}
(\bfn^{(\underline{l}:\overline{l}-1)}  | n^{(\overline{l})} ) .
\label{eq:ParallelDecompositionConditional} 
  \end{align}  
  
Parallel modules often share the same input neurons.  In such cases, 
a DuPlication (DP)  layer is introduced, which duplicates the neurons $\{n^{(\underline{l}-1)}\}$ at layer $\underline{l}-1$ to the input layer $\underline{l}$ of each parallel module, thereby making them non-overlapping, as shown in \Cref{fig:A.Modules} (right).

\paragraph{Hierarchical Layer Structure}
For a series connection of modules, it is often convenient to describe the relevance structure in a marginalized form.  Specifically, the intermediate layers within each module are marginalized out, leaving only the input and output layers of each module explicit. 
For example, we describe the relevance of a network consisting of two modules, Module 1 (M1) and Module 2 (M2), connected in series as 
\begin{align}
R(n^{(0)}, n^{(l)}| n^{(L)}) 
&=
R_{\mathrm{M1}}(n^{(0)}| n^{(l)})\, R_{\mathrm{M2}}(n^{(l)}| n^{(L)} ).
\label{eq:SeriesDecompositionConditionalMarginalized} 
\end{align}
Furthermore, because referencing global layer indices is cumbersome and provides little benefit, we assign layer indices \emph{separately} within each hierarchy and each module.  Under this convention, the top-level structure \eqref{eq:SeriesDecompositionConditionalMarginalized} is represented as a two-layer \emph{module-wise} network,
\begin{align}
R(n^{(0)}, n^{(1)}| n^{(2)}) 
&=
R_{\mathrm{M1}}(n^{(0)}| n^{(1)})\, R_{\mathrm{M2}}(n^{(1)}| n^{(2)} ),
\notag
\end{align}
and the relevance of each module is defined locally using its own layer indices, i.e.,
\begin{align}
R_{\mathrm{M1}}(\bfn^{(0:l-1)} | n^{(l)})
&= \textstyle \prod_{l'=1}^l R_{\mathrm{M1}}(n^{(l'-1)}| n^{(l')}),
\notag\\
 R_{\mathrm{M2}}(\bfn^{(0:L-l-1)}| n^{(L-l)})
 &= \textstyle \prod_{l'=1}^{L-l} R_{\mathrm{M2}}(n^{(l'-1)}| n^{(l')}).
\notag
\end{align}
This 
implicitly defines the full joint relevance of the entire network with the global layer indices:
\begin{align}
R(\bfn^{(0:L-1)}| n^{(L)}) 
&=
R_{\mathrm{M1}}(\bfn^{(0:l-1)}| n^{(l)}) R_{\mathrm{M2}}(\bfn^{(l:L-1)}| n^{(L)} ).
\notag
  \end{align}  
  
Since the LRP algorithm is formulated as a sum-product message passing \citep{NRMpaper} algorithm, relevance propagation can also be carried out locally within each module.

\subsection{Example Application to LSTM}

\begin{figure}[t]
    \centering
    \includegraphics[width=0.6\linewidth]{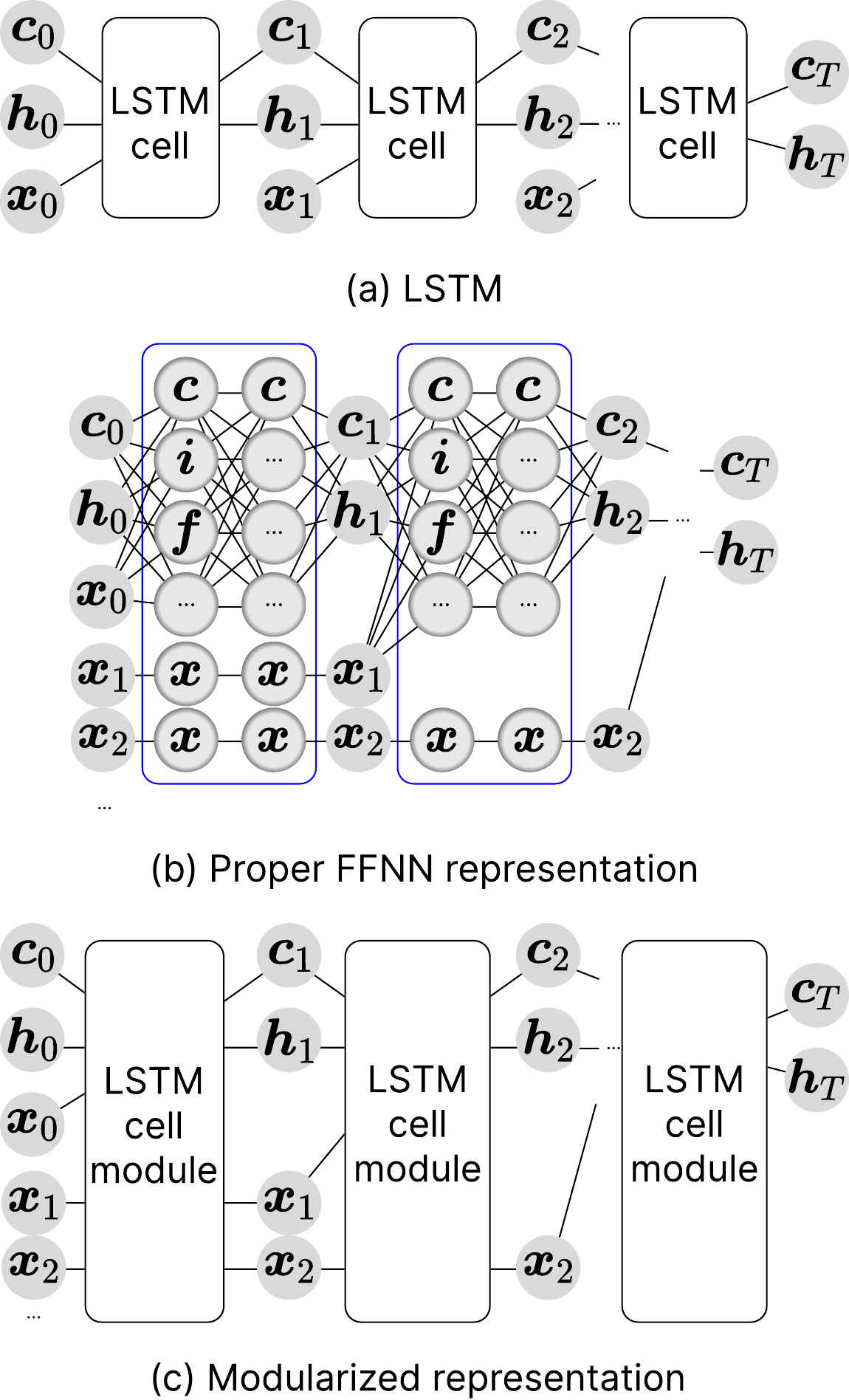}
    \caption{
        Modularization applied to an LSTM network (a). 
        Its proper FFNN  representation (b) consists of LSTM cell modules connected in series (c), for which the top-level relevance is decomposed as Eq.\eqref{eq:lstm_conditioned_rel_product}. 
    }
    \label{fig:lstm}
\end{figure}

Here we apply our modular decomposition procedure to LSTM \cite{DBLP:journals/neco/HochreiterS97} (\Cref{fig:lstm}a) as an illustrative example.  LSTM processes an input sequence $\{\bfx_t\}_{t=0}^{T-1}$ by recursively applying the LSTM cell,
\begin{align}
\bfi_t &= \sigma(\bfW_i \bfx_{t-1} + \bfU_i \bfh_{t-1}), \label{eq:LSTM.F.first}\\
\bff_t &= \sigma(\bfW_f \bfx_{t-1} + \bfU_f \bfh_{t-1}), 
\label{eq:LSTM.F.second}\\
\bfo_t &= \sigma(\bfW_o \bfx_{t-1} + \bfU_o \bfh_{t-1}), 
\label{eq:LSTM.F.third}\\
\tilde{\bfc}_t &= \tanh(\bfW_c \bfx_{t-1} + \bfU_c \bfh_{t-1}), \label{eq:LSTM.F.forth}\\
\bfc_t &= \bff_t \odot \bfc_{t-1} + \bfi_t \odot \tilde{\bfc}_t, \label{eq:LSTM.F.fifth}\\
\bfh_t &= \bfo_t \odot \tanh(\bfc_t), 
\label{eq:LSTM.F.sixth}
\end{align}
 at each time step for $t = 1, \ldots, T$,
where $\sigma(\cdot)$ and $\odot$ denote the sigmoid function and  element-wise multiplication, respectivey. Here we omitted bias terms for simplicity.

To apply the NRM framework, we first convert the network into a proper FFNN following Xiong et al.~\cite{NRMpaper}: the recurrent architecture is unfolded, intermediate inputs and outputs are copied to the input and output layers, respectively, and skip connections are eliminated by duplicating neurons across the skipped intermediate layers (\Cref{fig:lstm}b). 

LSTM can be seen as a series connection of LSTM cell modules, where each module takes as inputs
the previous hidden and cell states $(\bfh_{t-1}, \bfc_{t-1})$ together with the input sequence $\{\bfx_{t'}\}$ for $t' \geq t-1$, and outputs the current hidden and cell states $(\bfh_t,\bfc_t)$ together with the input sequence $\{\bfx_{t'}\}$ for $t' \geq t$ (\Cref{fig:lstm}c).
Therefore, following the modular decomposition rule in Eq.~\eqref{eq:SeriesDecompositionConditional} and the hierarchical layer indexing convention, the top-level relevance of LSTM is decomposed as
\begin{align}
    R_{\mathrm{LSTM}}(\bfn^{(0: T-1)} | n^{(T)}) = \prod_{t=1}^{T}R_{\mathrm{LSTM-Cell}}(n^{(t-1)}| n^{(t)}),
    \label{eq:lstm_conditioned_rel_product}
\end{align}
where $n^{(\cdot)}$ specifies a neuron in each (local) layer as
\begin{align}
n^{(t)} \in \mcS^{(t)}( \{ \bfh_{t}, \bfc_{t}, \{\bfx_{t'}\}_{t'=t}^{T-1} \}).      \label{eq:GlobalDecompositionNeuronsLSTM}
\end{align}
Here, $\mcS^{(l)}({{\mcX}})$ denotes the whole set of neuron indices in the $l$-th layer that represent a set of variables $\mcX$.
Defining the local relevance $R_{\mathrm{LSTM-Cell}}(\cdot)$ for each module, which is detailed in \ref{app:lstm}, completes the entire relevance definition.

This example, together with another example of modularization involving parallel connections in  \ref{sec:A.ExampleModularization}, highlights the advantages of modularization: the relevance can be defined locally and assembled hierarchically into the entire structure. The same principle is applied in \Cref{sec:ExplainETGNN} to handle the considerably more complex architecture of ETGNNs.

\section{Explaining Event-based Temporal Graph Neural Networks (ETGNNs)}
\label{sec:ExplainETGNN}

We apply NRM and explain one of the most general and popular variants of ETGNN 
proposed by Rossi et al.~\cite{DBLP:journals/corr/abs-2006-10637}.
Events serve as the primary inputs of ETGNNs \citep{DBLP:journals/corr/abs-2006-10637,DBLP:conf/sigir/0001GRTY20}, making event-level explanation a natural choice for explaining the model. The challenge, however, lies in defining a meaningful attribution quantity (relevance score) for each event. Here, we propose a principled approach to defining and computing such scores within the NRM framework.
We first briefly introduce temporal graphs and events, and describe the forward process of ETGNNs.
Then, we give an explicit definition of relevance structure for the entire model, and propose our novel event relevance.

\subsection{Temporal Graphs and Events}
\label{sec:TemporalGraph}

A temporal graph is specified by an initial graph $\mathcal{G}_0 = (\mathcal{V}, \mathcal{A}_0)$ with a set $\mathcal{V}$ of nodes and a set $\mathcal{A}_0$ of edges, 
together with a set $\mathcal{E}$ of events with time stamps.
The temporal graph $\mathcal{G}_t = (\mathcal{V}, \mathcal{A}_t)$ evolves at the time when each event occurs with addition  of the edge between the associated nodes\footnote{Removal of edge or node-level event are also possible. We focus on addition-only definition for simplicity, but our approach can straightforwardly be extended to cover other cases.}. 
Each event $e \in \mathcal{E}$ is associated with the origin node $v_{\mathrm{o}}^{(e)}$, the destination node $v_{\mathrm{d}}^{(e)}$, the event encoding $\bftau^{(e)}$, and the event occurring time $t^{(e)} >0$.  
We define the \emph{event feature} as
\begin{align}
\bfepsilon^{(e)}(t) = (\bftau^{(e)}, \bfphi^{}(|t - t^{(e)}|))  \in \mathbb{R}^{D_{\mathrm{E}}},
\label{eq:EventEmbedding}
\end{align}
where $\bfphi(\cdot)$ is a time encoding function and $t$ is the time when the feature is referred to (e.g., inference time).
Each event affects the graph at the time $t^{(e)}$ by connecting the origin node $v_{\mathrm{o}}^{(e)}$ and the destination node $v_{\mathrm{d}}^{(e)}$, and updating their node features.

\subsection{Forward Process of ETGNNs}
\label{sec:ETGNNForwardProcess}

\begin{figure}[!t]
    \centering
    \begin{subfigure}{0.95\linewidth}
        \centering
        \includegraphics[width=\linewidth]{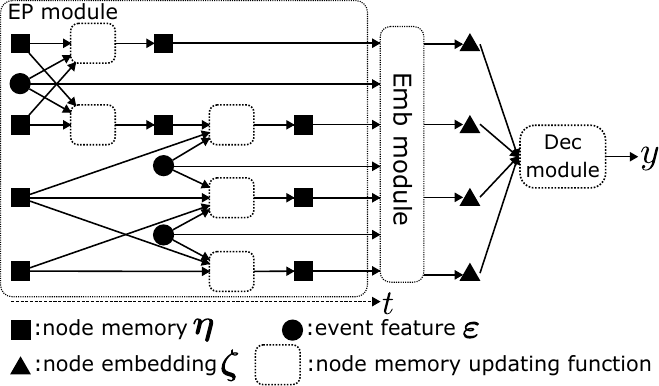}
        \caption{Forward computation of ETGNN. Node memories are initialized at $t=0$, and updated in the Event Processing (EP) module. 
    After the last training time point $t = T$, the Embedding (Emb) module finalizes the node memories into the final node embeddings, which are used in the Decoding (Dec) module for downstream prediction at $t = T^* \ge T$.}
        \label{fig:tgn_whole_model}
    \end{subfigure}
    \vskip 0.5em
    \begin{subfigure}{0.95\linewidth}
        \centering
        \includegraphics[width=\linewidth]{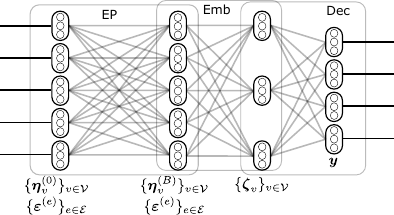}
        \caption{ETGNN as a \emph{proper} FFNN with three modules. Each module consists of many layers.}
        \label{fig:ETGNNasProperFFNN}
    \end{subfigure}
    \caption{Overview of ETGNN: (a) forward computation process, and its (b) equivalent proper FFNN view.}
    \label{fig:ETGNN_combined}
\end{figure}

An ETGNN, illustrated in \Cref{fig:tgn_whole_model}, 
consists of three modules, an \emph{Event Processing} (EP) module, an \emph{Embedding} (Emb) model, and 
a \emph{Decoding} (Dec) module.
In the EP module, each node $v \in \mathcal{V}$ is assigned a dynamic node feature $\bfeta_v(t) \in \mathbb{R}^{D_{\mathrm{V}}} $, called \emph{node memory},
which is initialized at $t = 0$ and evolves over time
along with the temporal graph connectivity.
After all events up to the time $t=T$ are processed,
the Emb module refines the node embeddings for the inference at $t = T^* (\ge T)$, based on the node memories and the latest graph $\mathcal{G}_{T}$.
The resulting node embeddings are used by the Dec module for downstream tasks.

\paragraph{Updating Node Memories}
In the EP module, events are processed batch-wise in the order of the event occurrence time,
and, in each batch process,
node memories 
$\bfeta_v (t)$ 
are updated with forward message exchanges between the origin and the destination nodes.
Let $\{\mcE_b\}_{b=1}^B = \mcE$ be the event set decomposition into $B$ batches.
Then, 
each node memory function is a step function with at most $B+1$ different values:
\begin{align}
\bfeta_v (t) = \bfeta_v^{(b)}  \quad \mbox{ for } \quad \overline{t}_{b} \leq t <  \overline{t}_{b+1}, 
\quad 
b = 0, \ldots, B,
\notag
\end{align}
where
$\overline{t}_{b} = \max_{e \in \mcE_b} t^{(e)}$ 
for $b = 1, \ldots, B$
is the time that the last event in the batch $\mcE_b$ occurred,
and we define 
$\overline{t}_{0} = 0$, $\overline{t}_{B} = T$, and $\overline{t}_{B+1} = \infty$.
Starting from the initial node memories
$\bfeta_v^{(0)}$, the following forward process is performed for the batches $b = 1, \ldots, B$. 

For each event $e \in \mcE_b $ in the $b$-th batch, 
we compute \emph{messages} 
\begin{align}
        \bfm_{v_1}^{(e)} &=
      \text{Message}_{\text{o}}(\bfeta_{v_1}^{(b-1)}, \bfeta_{v_2}^{(b-1)} , \bfepsilon^{(e)}(t^{-}_{v_1})  ), 
      \label{eq:tgn_message_func_org}\\
      \bfm_{v_2}^{(e)} &= \text{Message}_{\text{d}}(\bfeta_{v_2}^{(b-1)}, \bfeta_{v_1}^{(b-1)} , \bfepsilon^{(e)}(t^{-}_{v_2})  ),
    \label{eq:tgn_message_func_dest}
\end{align}
for the origin $v_1 = v_{\mathrm{o}}^{(e)}$ and destination $v_2 = v_d^{(e)}$ nodes.
Here,
$t^{-}_{v}$ is the latest update time of the node $v$, and   
$\text{Message}_{\text{o}}(\cdot)$ and 
 $\text{Message}_{\text{d}}(\cdot)$ are forward message functions, typically  implemented with MLPs.

After all messages invoked by all events in the batch $\mcE_b$ are generated, they are aggregated for each node, with which the node memory is updated.  Specifically, for each node $v \in \mcV$ associated with at least one event $e\in \mcE_b$ in the batch, we compute
\begin{align}
        \widehat{\bfm}_v &= \text{Aggregate}(\{\bfm_v^{(e)} \}_{e\in\mcE_b }), 
       \label{eq:tgn_aggregate_func}\\
   \bfeta_{v}^{(b)} &= \text{Update}(\widehat{\bfm}_v, \bfeta_{v}^{(b-1)} ),   
   \label{eq:tgn_update_func}
\end{align}
where 
$\text{Aggregate}(\cdot)$
and $\text{Update}(\cdot)$ are aggregate and update functions, respectively.
Typical choices for the aggregate function include sum/average over the messages,
and
a usual choice for the update function is 
a Gated Recurrent Unit (GRU) cell
\citep{DBLP:conf/emnlp/ChoMGBBSB14}.
The memory of the nodes that do not receive message 
is unchanged, i.e., $\bfeta_{v}^{(b)} = \bfeta_{v}^{(b-1)} $.

\paragraph{Inference Details} 
Before making a prediction at time $t = T^* > T$, the Emb module refines the node memories to the final node embeddings:
\begin{align}
    \bfzeta_v &= \text{Emb}(v, \{\bfeta_{v'}^{(B)}\}_{v' \in \mathcal{V}}, \mathcal{G}_{T^*}  ),
    \label{eq:tgn_embedding_func}
\end{align}
where $\text{Emb}(\cdot)$ is the embedding function, which can be simply a copy of the latest memory 
$\bfeta_{v}^{(B)}$, or a multilayer attention mechanism \citep{DBLP:conf/iclr/XuRKKA20} that aggregates the memories of the neighboring nodes.

With the final node embeddings, the downstream node-, edge- and graph-level  predictions are performed as
\begin{align}
    \bfy_{\mathrm{node}} &= 
            \text{Dec}_{\mathrm{node}}(\bfzeta_{v}), \\
        \bfy_{\mathrm{edge}} &= \text{Dec}_{\mathrm{edge}}(\bfzeta_{v_{1}}, \bfzeta_{v_{2}}), \\
        \bfy_{\mathrm{graph}} &= \text{Dec}_{\mathrm{graph}}(\{\bfzeta_{v}\}_{v \in \mathcal{V}}), 
        \label{eq:tgn_predict_func}
\end{align}
respectively,
where $\text{Dec}_{\;\text{---}}(\cdot)$ is a downstream decoding module, which is typically an MLP.
Here, the node $v$ and edge $(v_1, v_2)$
are specified by the user.

\subsection{Relevance Definition for ETGNNs}
\label{sec:ETGNN.JointRelevanceDefinition}

After converting the ETGNN to a proper FFNN (following the procedure in Xiong et al.~\cite{NRMpaper} and our extension of modularization),
the entire network (\Cref{fig:tgn_whole_model})
can be seen as a series connection of the EP, Emb, and Dec modules (see \Cref{fig:ETGNNasProperFFNN}). 
The EP module takes as inputs the initial node memories $\{\bfeta_v^{(0)}\}_{v\in \mcV}$
and the event embeddings $\{\bfepsilon^{(e)}\}_{e \in \mcE}$,
   and provides as outputs the last node memories $\{\bfeta_v^{(B)}\}_{v\in \mcV}$  and the event embeddings $\{\bfepsilon^{(e)}\}_{e \in \mcE}$.  Note that the event embeddings $\{\bfepsilon^{(e)}\}_{e \in \mcE}$ in the outputs are exact copies of those in the inputs---which is required for the network to be a proper FFNN.
      The Emb module takes those outputs, and provides the final node embeddings $\{\bfzeta_v\}_{v \in \mcV}$.  The Dec module then uses the node embeddings for predicting the final output $\bfy$.  
      
   Let us summarize the variables introduced above as $\vec{\mcV}^{(b)} = \{\bfeta_v^{(b)}\}_{v\in \mcV}$,
   $\vec{\mcE} = \{\bfepsilon^{(e)}\}_{e \in \mcE}$, $\vec{\mcZ} = \{\bfzeta_v\}_{v \in \mcV}$, and $\vec{\mcY} = \{\bfy\}$.
Following our modularization procedure with hierarchical layer indexing (see  \Cref{sec:Modularization.MOdularization}),
the top-level relevance structure for the entire ETGNN is decomposed as a three-layer modular-wise network:
\begin{align}
&R_{\mathrm{ETGNN}}(n^{(0)} , n^{(1)} , n^{(2)} | n^{(3)})
= \notag \\
&\quad   R_{\mathrm{EP}}(n^{(0)} | n^{(1)} )
 R_{\mathrm{Emb}}( n^{(1)} | n^{(2)} )
 R_{\mathrm{Dec}}(n^{(2)} | n^{(3)}).
      \label{eq:GlobalDecomposition}
\end{align}
Here $n^{(\cdot)}$ 
specifies a single neuron from the set of neurons that represent the variables at the corresponding layer, i.e.,
\begin{align}
&n^{(0)} \in \mcS^{(0)}({\vec{\mcV}^{(0)} \cup \vec{\mcE}}), \,
n^{(1)} \in \mcS^{(1)}({\vec{\mcV}^{(B)} \cup \vec{\mcE}}), \, \notag\\
&n^{(2)} \in \mcS^{(2)}({\vec{\mcZ}}), \,
n^{(3)} \in \mcS^{(3)}({\vec{\mcY}}).
     \label{eq:GlobalDecompositionNeurons}
\end{align}
For example, $n^{(1)}$ specifies a neuron from those for ${\vec{\mcV}^{(B)} \cup \vec{\mcE}}$, which are the output of the EP module, and the input of the Emb module.

Each factor in Eq.\eqref{eq:GlobalDecomposition} should be further decomposed into relevances of submodules.  Iterating decompositions hierarchically, we obtain the definition of the full joint relevances, i.e., the relevance of all individual walks.
This procedure is detailed in 
\ref{sec:A.ETGNN.JointRelevanceDefinition}.

\subsection{Event Relevance}
\label{sec:def_evt_rel}

We propose a novel definition of event relevance (ER), considering three types of information flow that are associated with each event.  
We further extend the idea to higher-order joint ER that focuses on the intersection of the information flows associated with multiple events. In addition, we propose the Grad$\times$Input \citep{shrikumar2017learning, ancona2018towards} counterpart that approximates our ER definition.

\begin{figure}[t]
    \centering
    \includegraphics[width=\linewidth]{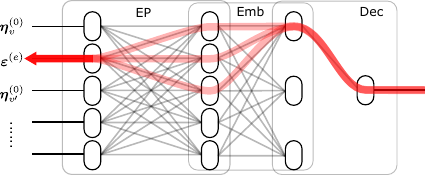}
    \caption{
    Illustration of
ER-feat for the event $e$, explaining the node-level prediction of node $v$.
    }
    \label{fig:ER-feat-simple}
\end{figure}

\begin{figure}[t]
    \centering
  \includegraphics[width=\linewidth]{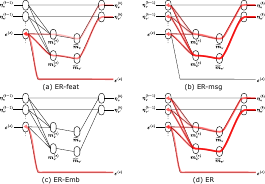}
    \caption{
    Illustration of four ER definitions.
    (a) ER-feat is the relevance of all walks arriving at the event feature $\boldsymbol{\varepsilon}^{(e)}$.
  (b)   ER-msg is the relevance of all walks passing through one of the two messages between nodes $v,v'$ induced by the event $e$. 
  (c)   ER-Emb extracts only the relevance received in the Emb module.
   (d)  ER is the relevance of the union of the walks for ER-feat, ER-msg, and ER-Emb, which is equal to    
   the sum of ER-feat and ER-msg.
    Note that the neurons over all layers  outside this figure, except the output layer, are marginalized. 
     }
    \label{fig:ER-msg-simple}
\end{figure}

 \paragraph{Event Relevance through Feature (ER-feat)}  
 A simple way to define the relevance of an event $e \in \mcE$ is to count the relevance of all walks that arrive at the  input neurons that represent the event feature $\bfepsilon^{(e)}$---concatenation of the event type and the time embeddings---defined in Eq.\eqref{eq:EventEmbedding}:
\begin{align}
\textrm{ER-feat}_{\; e|n^{(L)}}
\equiv
 R(n^{(0)} \in \mcS(\{\bfepsilon^{(e)}\}) | n^{(L)}),
 \label{eq:ERemb}
\end{align}
where $n^{(0)}$ and $n^{(L)}$
specify neurons at the input $l=0$ and the output $l=L$ layers of the entire ETGNN.
This relevance collects all relevances that the event feature $\bfepsilon^{(e)}$ receives both in the EP and Emb modules. 
\Cref{fig:ER-feat-simple} and \Cref{fig:ER-msg-simple} (a) illustrate this quantity as the corresponding set of walks.

\paragraph{Event Relevance through Messages (ER-msg)}
To account for how an event affects the memory updates of the associated nodes,
ER-msg collects all relevances passing through the neurons representing one of the two messages $\{\bfm^{(e)}_{v_\mathrm{o}^{(e)}}, \bfm^{(e)}_{v_{\mathrm{d}}^{(e)}}\}$ induced by the event $e \in \mcE$ in the EP module:
\begin{align}
&\textrm{ER-msg}_{\; e|n^{(L)}}
\equiv \notag \\& \qquad  \textstyle
R\left(n^{(l_{\mathrm{msg}(e)})} \in 
\mcS({\{\bfm_{v_{\mathrm{o}}^{(e)}}^{(e)}, \bfm_{v_{\mathrm{d}}^{(e)}}^{(e)}\}})
| n^{(L)}\right).
 \label{eq:ERMsg}
\end{align}
Here, $l_{\mathrm{msg}(e)}$ is the (globally indexed) layer
where the neurons for the message variables appear in the EP module.
\Cref{fig:ER-msg-simple} (b) illustrates ER-msg, where the neuron variables in 
 all other layers except the output layer are marginalized.

\paragraph{Event Relevance through Emb Module (ER-Emb)}
The third definition of event relevance only considers the relevance that the event feature receives in the Emb module.
Focusing on the same layer $l_{\mathrm{msg}(e)}$ as for ER-msg, it can be evaluated as
\begin{align}
\textrm{ER-Emb}_{\; e|n^{(L)}}
\equiv \textstyle
R\left(n^{(l_{\mathrm{msg}(e)})} \in 
\mcS(\{\bfepsilon^{(e)}\})
| n^{(L)}\right), 
    \label{eq:ER-Emb}
    \end{align}
as illustrated in \Cref{fig:ER-msg-simple} (c).
Note that we define ER-Emb not as a promising ER definition, but as an NRM counterpart of the existing XAI methods, e.g., TGNNExplainer, that ignore the entire process within the EP module.
Indeed, ER-Emb performs as poorly as the existing methods, as seen in \Cref{sec:Experiment}.

\paragraph{Event Relevance (ER)}

Our proposed ER is the relevance of the union of the walks considered above, which actually amounts to the sum of ER-msg and ER-Emb (see \Cref{fig:ER-msg-simple} (b)-(d)):
\begin{align}
&\textrm{ER}_{\; e|n^{(L)}}
\equiv \notag \\& \quad \textstyle
R(n^{(l_{\mathrm{msg}(e)})} \in 
\mcS(\{\bfm_{v_{\mathrm{o}}^{(e)}}^{(e)}, \bfm_{v_{\mathrm{d}}^{(e)}}^{(e)}, {\bfepsilon^{(e)}}\})
| n^{(L)}).
\label{eq:ER-all}
    \end{align}

\paragraph{Joint Relevance of Multiple Events}

\begin{figure}
    \centering
    \includegraphics[width=0.9\linewidth]{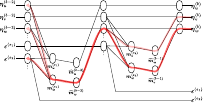}
    \caption{Joint relevance of events $e_1$ and $e_2$, where $e_1$ is associated with nodes $u$ and $w$ and $e_2$ is with nodes $v$ and $w$.}
    \label{fig:joint_two_event_rel_simple}
\end{figure}

Our ER definition can be naturally generalized to joint relevance of multiple events.  Specifically, we define the relevance for a set of events, $\{e_1, \cdots, e_K\}$, as 
\begin{align}
&\textrm{ER}_{\; e_1, \ldots, e_K|n^{(L)}}
\equiv \notag \\& \;\;  \textstyle
R\bigg(n^{(l_{\mathrm{msg}(e_1)})} \in \mcS({\{\bfm_{v_{\mathrm{o}}^{(e_1)}}^{(e_1)}, \bfm_{v_{\mathrm{d}}^{(e_1)}}^{(e_1)}\}}),
\ldots, \notag\\
&\; \; \; \; \; \; n^{(l_{\mathrm{msg}(e_K)})} \in \mcS({\{\bfm_{v_{\mathrm{o}}^{(e_K)}}^{(e_K)}, \bfm_{v_{\mathrm{d}}^{(e_K)}}^{(e_K)}\})}
| n^{(L)} \bigg).
 \label{eq:JointERMsg}
\end{align}
This quantity focuses on the information flow passing through one of the two message variables associated with each specified event, and thus captures higher-order interactions between the events. An example of joint relevance of two events is illustrated in \Cref{fig:joint_two_event_rel_simple}.

\paragraph{Grad$\times$Input-msg} 

Grad$\times$Input \citep{shrikumar2017learning, ancona2018towards} is a simple explanation method, which is known to be equivalent to LRP-0 \citep{bach2015pixel} in FFNNs with ReLU activations.
We propose to use Grad$\times$Input applied to the messages induced by the event $e\in \mcE$, as an approximation to our ER-msg.

\section{Empirical Evaluation of Explaining ETGNNs}
\label{sec:Experiment}

In this section, we conduct
qualitative and quantitative experiments with synthetic and real-world temporal graph datasets, and compare our methods with baselines.\footnote{The implementation will be made publicly available upon publication.} Specifically, we consider two artificial and one real-world datasets:
a simulated infection network (\Cref{sec:infection_experiment}), simulated attacker subgraphs in a social network (\Cref{sec:attacker_experiment}),
and a real-world political events network (\Cref{sec:icews_experiment}).

The model configurations and performances are listed in \ref{sec:model_info}. 
Details of the datasets are described in each corresponding subsection.

Across all datasets, in addition to qualitatively interpreting the results, we quantitatively evaluate the performance of our explanation method and compare it against baseline methods. Specifically, we use both ground-truth-based and perturbation-based metrics \cite{samek2017eval_xai}: For synthetic datasets with ground-truth labels, we measure the precision and recall of the top-$k$ most relevant features, while for real datasets where ground-truth labels are unavailable, we instead evaluate pruning and activation curves, which assess whether the attributed relevance to each feature is a good estimate of the decrease in model output when that feature is removed from the input. More details on the evaluation metrics can be found in \ref{sec:quant_eval_method}.

\begin{figure*}[t]
    \centering
    \includegraphics[width=\linewidth]{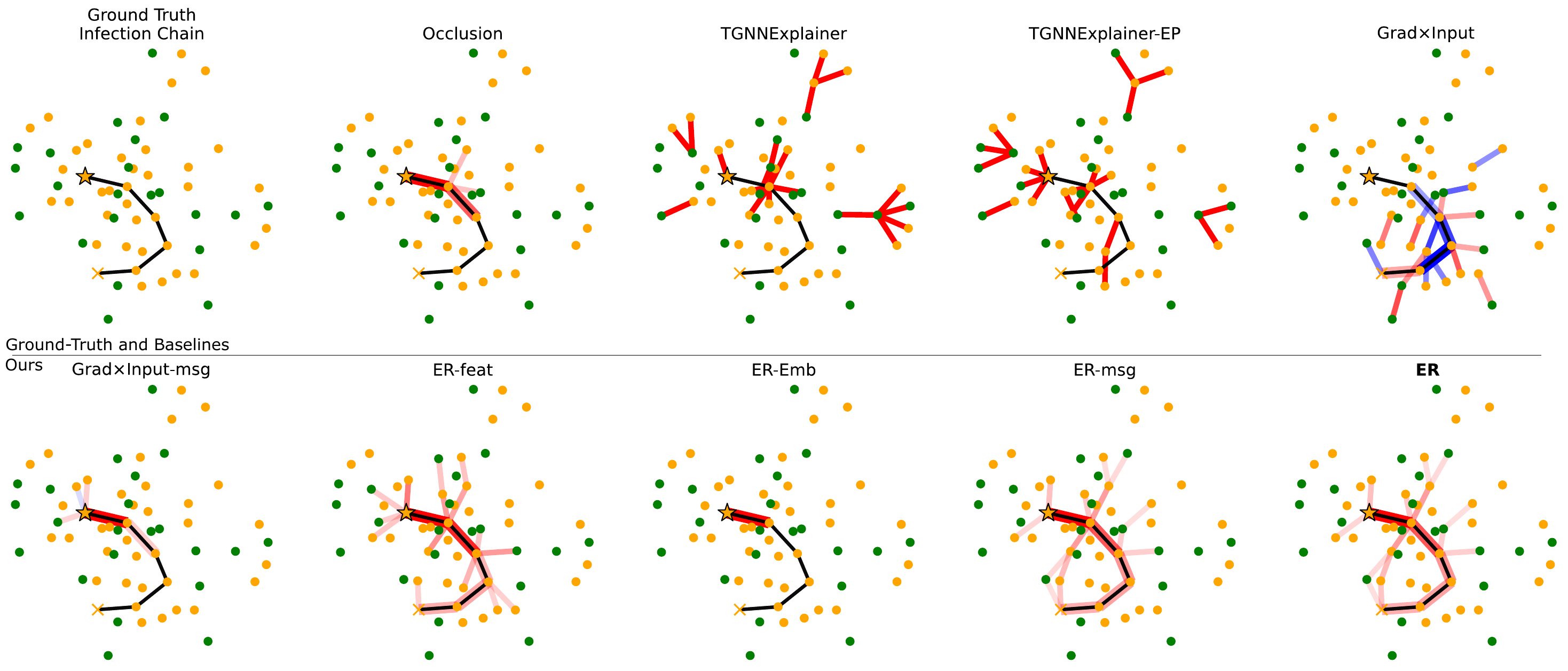}
    \vskip -0.9em
    \caption{  Top-20 most relevant event edges for node prediction for an example episode in the Infection dataset. 
    Nodes are depicted as circles except special ones: the crosses are the nodes initially infected, and the star is the node to be predicted.
    The yellow and green colors indicate that the node is infected and not-infected, respectively, at the inference time.
    A ground-truth chain is shown as the connected black line segments. 
    Our methods, 
    ER-feat, ER-msg, and ER, successfully identify the chain as positively relevant, while the others fail. 
    }
    \label{fig:infection_baselines_cmp}
\end{figure*}

\paragraph{Baseline Methods}

As discussed in \Cref{sec:BG.TGNNExp},
existing explanation methods for ETGNN, including
TGNNExplainer \citep{DBLP:conf/iclr/XiaLS0D0023} and TempME \citep{DBLP:conf/nips/ChenY23}, ignore the influence of events to the node memories in the EP module,
and only focus on how the event features influence the final node representations in the Emb module, and thus to the final predictions in the Dec module.
As baselines, we evaluate TGNNExplainer and its adapted version.
The original \textbf{TGNNExplainer} is evaluated only in the experiments where the Emb module is not identity,
while the adapted version, named \textbf{TGNNExplainer-EP}, is evaluated in all experiments.
The latter also accounts for the information flow in the EP module by perturbing the node memories when calculating the reward values in MCTS.%
\footnote{It is not straightforwared to adjust the MLP accelerator of TGNNExplainer such that it masks events in the EP module that involves GRU cells.  Therefore, we do not use the MLP accelerator in TGNNExplainer-EP.  This does not affect performance, because it is only for accelerating computation.
}
We also compare our method against Occlusion and Grad$\times$Input, which are
model agnostic and therefore generally applicable independent of the model architecture.
\textbf{Occlusion} \citep{BLUCHER2022103774} defines  the relevance of an event $e$ as the change in prediction 
when the event is masked out.
\textbf{Grad$\times$Input} \citep{shrikumar2017learning, ancona2018towards} defines the relevance score of an event as the product of the gradient and the input for the event embedding $\bfepsilon^{(e)}$. 

\medskip

In all ER methods, we use LRP-$\gamma$ rule with $\gamma=0.1$ for linear layers, and LRP-all rule for element-wise multiplication in GRU and attention mechanism. 
We add a small term $\varepsilon = 10^{-6}$ to numerator of LRP-$\gamma$ rule 
 to ensure numerical stability.

We summarize all average metric scores over top-$k$ events in \Cref{tab:experiment_results}.

\begin{table*}[t]
\centering
\begin{tabular}{l|l|llll|ll}
\textbf{Dataset} & Infection      & \multicolumn{4}{l|}{Attacker}                                & \multicolumn{2}{l}{ICEWS18}     \\ \hline
\textbf{Metric}  & Recall-chain   & Precision      & Recall         & Pruning        & Activation     & Pruning        & Activation     \\ \hline
ER (ours)              & \textbf{0.844} & \textbf{0.874} & \textbf{0.535} & {0.843} & \textbf{0.565} & \textbf{0.098} & 0.173          \\
ER-feat (ours)          & 0.563          & 0.726          & 0.415          & \textbf{0.845}          & 0.206          & 0.092          & 0.170          \\
ER-msg (ours)           & \textbf{0.844} & \textbf{0.874} & \textbf{0.535} & {0.843} & \textbf{0.565}   & 0.065          & 0.180          \\
ER-Emb (ours)           & 0.104          & -              & -              & -              & -              & 0.090          & 0.174          \\
G$\times$I-msg (ours)   & 0.183          & 0.180          & 0.104          & 0.166          & 0.020          & 0.023          & 0.183          \\
\hline
G$\times$I       & 0.125          & 0.290          & 0.171          & 0.255          & 0.006          & 0.003          & \textbf{0.198} \\
Occlusion        & 0.352          & 0.714          & 0.450          & 0.698          & 0.443          & -              & -              \\
TGNNExp          & 0.055          & 0.237          & 0.142          & 0.336          & 0.025          & -0.021         & 0.077          \\
TGNNExp-EP       & 0.021          & 0.265          & 0.171          & 0.316          & 0.188          & -0.023         & 0.066         
\end{tabular}
\caption{Average Scores of different metrics over top-$k$ events. 
For the Attacker experiment, 
the model uses Identity Emb module, and thus ER-msg is equal to ER and ER-Emb is not applicable. For ICEWS18, the runtime of Occlusion is infeasible. We highlight the best scores in bold.
Detailed results including precision, reall, pruning, and activation curves can be found in \ref{app:qualitative_experiment}. 
}
\label{tab:experiment_results}
\end{table*}

\subsection{Explaining Infection Network Predictions}
\label{sec:infection_experiment}

We simulate an infectious disease spreading within a pool of 500 people (nodes). In each time step, 20 events (interactions between a pair of nodes) occur. In each interaction, infection occurs if one is sick and the other is not, with probability $\alpha$ that depends on a binary event feature---%
$\alpha =0.1$
if both wear masks and $\alpha = 0.9$ otherwise.
For simplicity, we assume that the disease is not recoverable. 
Starting from 5 initial infected nodes,
we simulate the spreading process for 1000 times.

The model is trained on a node-level binary classification task, where the positive class means that the node is infected at the prediction time. We generated 100 graphs and used 80\% for training and 20\% for testing. Note that for this dataset, the oracle (highest possible) accuracy is $\approx 84\%$, which was computed by simulating 
the probability for each node to be infected at the prediction time.
 We also identified the ground-truth set
of infection chains, based on their frequency in the simulation.  Specifically, we collect as the ground-truth chains all those with probability larger than $5\%$.
We trained a model with a single layer attention Emb module, and a model with Identity Emb module, whose prediction accuracies are $81.6\%$ and $80.7\%$, respectively.

\subsubsection{Qualitative Visualization}
\label{sec:infection_qualitative}

For qualitative evaluation, we use the model with the attention Emb module.
\Cref{fig:infection_baselines_cmp} shows an explanation of node prediction for an example temporal graph.  The infection of the node marked as yellow star is predicted, for which each explanation method provides the top-20 most relevant events.   In this example, the most probable infection chain is the one shown as black connected edges.  ER-feat, ER-msg, and ER identified this entire ground-truth infection chain within the top-$20$ relevant events, while the baseline methods, as well as ER-Emb that ignores the entire process in the EP module, at best only find a few last steps.  
This result clearly shows that accounting for the information flow in the EP module is essential for good explanation.
Note that other ground-truth chains with lower probability are not identified by any method.  This is reasonable because the model can predict well if one infection chain is recognized.

\subsubsection{Quantitative Evaluation}
\label{sec:infection_quantitative}
Since the ground-truth infection chains are available, we evaluate \emph{Recall-chain}---a metric evaluates whether the top-$k$ events can form a chain that connects the initial infected person to the target person---defined in Eq.\eqref{eq:recall-chain} in \ref{sec:quant_eval_method}.

The second column in \Cref{tab:experiment_results}
reports on the average recall chain over the top-$k$ events (see \ref{app:qualitative_experiment} for the  recall chain curves).
We observe that ER methods (except ER-Emb) significantly outperform the baseline methods, quantitatively confirming the qualitative findings above. 
 Furthermore, ER-feat is significantly weaker than ER-msg and ER, 
  which supports our strategy of accounting for the full event-induced information flow, including pathways through event-associated message variables, for event relevance assessment.
 Note the difference between ER-feat and ER-msg, shown in \Cref{fig:ER-msg-simple} (a) and (b): although both account for walks going through one of the two associated messages, ER-feat counts only those arriving at the event feature in the input layer, while ER-msg counts those arriving at any neuron in the input layer.
 ER-Emb, as well as TGNNExplainer, performs poorly, indicating that the primary event-induced processes occur in the EP module, rather than in the Emb module.%
 \footnote{
We considered the possibility that the poor performance of TGNNExplainer stems from initializing the event coalition with the 25 closest two-hop events, as proposed in the original paper. However, we confirmed that its performance remained similarly poor even when different initial event coalitions are used.
 }

\subsubsection{Relevance of Longer Infection Chains}
\label{sec:partial_walk_infection}

\begin{figure}[t]
\centering
{
\small
\begin{tabular}{lccc}
\multicolumn{4}{l}{Event Relevances Sorted by Absolute Value}                                \\ 
\hline
\multicolumn{1}{l|}{\textbf{482\eventtime{5}1}} & \cellcolor{red!100}2.1054 &\multicolumn{2}{c}{\multirow{14}{*}{ \includegraphics[width=0.6\linewidth]{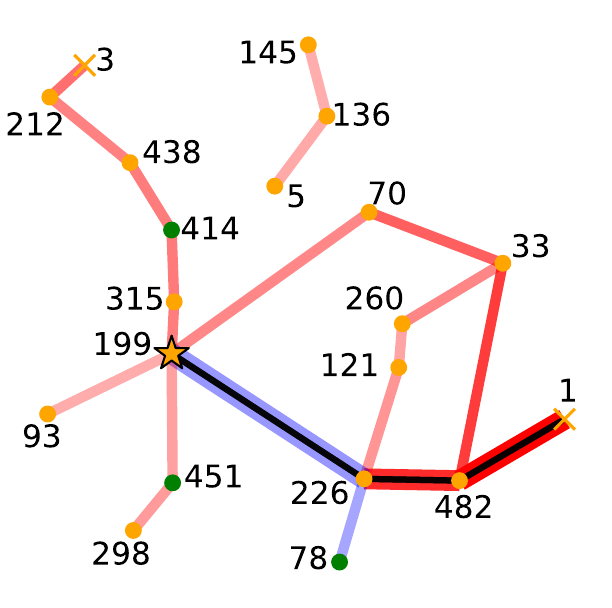}}}\\
\multicolumn{1}{l|}{\textbf{226\eventtime{39}482}} & \cellcolor{red!69}1.4617 \\
\multicolumn{1}{l|}{33\eventtime{15}482} & \cellcolor{red!59}1.2365 \\
\multicolumn{1}{l|}{70\eventtime{40}33} & \cellcolor{red!40}0.8336 \\
\multicolumn{1}{l|}{212\eventtime{2}3} & \cellcolor{red!32}0.6683 \\
\multicolumn{1}{l|}{414\eventtime{50}315} & \cellcolor{red!27}0.5749 \\
\multicolumn{1}{l|}{438\eventtime{43}414} & \cellcolor{red!27}0.5593 \\
\multicolumn{1}{l|}{315\eventtime{60}199} & \cellcolor{red!26}0.5458 \\
\multicolumn{1}{l|}{212\eventtime{7}438} & \cellcolor{red!24}0.5124 \\
\multicolumn{1}{l|}{260\eventtime{18}33} & \cellcolor{red!23}0.4740 \\
\multicolumn{1}{l|}{199\eventtime{60}70} & \cellcolor{red!22}0.4670 \\
\multicolumn{1}{l|}{226\eventtime{66}121} & \cellcolor{red!17}0.3654 \\
\multicolumn{1}{l|}{\textbf{226\eventtime{77}199}} & \cellcolor{blue!17}-0.3488 \\
\multicolumn{1}{l|}{...} & ... \\
\hline
\multicolumn{4}{l}{Top Joint  Relevance of 3 Events}                                                          \\
\hline
\multicolumn{3}{l|}{\textbf{482\eventtime{5}1, 226\eventtime{39}482, 226\eventtime{77}199}} & \cellcolor{red!53}1.1251 \\
\multicolumn{3}{l|}{33\eventtime{15}482, 70\eventtime{40}33, 199\eventtime{60}70} & \cellcolor{red!35}0.7277 \\
\multicolumn{3}{l|}{482\eventtime{5}1, 33\eventtime{15}482, 70\eventtime{40}33} & \cellcolor{red!28}0.5949 \\
\multicolumn{3}{l|}{...} & ...
\end{tabular}
}
\caption{Marginal and joint event relevances for predicting the infection of node 199 (marked with \textcolor{orange}{$\bigstar$}).
The ground-truth infection chain is shown as black lines, and 
the event between the nodes $a$ and $b$ at time $t$ is denoted as $a$\eventtime{t}$b$. 
Top: Marginal event scores by ER, which fails to identify the entire ground-truth chain as all positively relevant edges. 
Bottom: Joint event relevance for three-event sets, which successfully identify the ground-truth chain as the most relevant set.
}
\label{tab:joint_rel_infection_199}
\end{figure}

Although the experiments above showed superior performance of our ER methods to the baselines, collecting the top-$k$ events based on the marginal relevance of single edge event is not optimal for 
identifying an entire infection chain---which intrinsically requires to capture higher-order interactions. 
The joint event relevance \eqref{eq:JointERMsg}
is expected to capture such higher-order interactions between events.
\Cref{tab:joint_rel_infection_199} shows an example infection scenario, where the single event relevance by ER fails to identify the entire infection chain: Although two of the event edges in the ground-truth chain (black lines) are identified as positively (red) relevant, the other event edge is identified as negatively (blue) relevant.
On the other hand, 
as shown in the bottom of \Cref{tab:joint_rel_infection_199}, where the top-3 jointly relevant three-event sets are reported, the ground-truth chain is successfully identified as the most relevant set. 
This result highlights an advantage of our joint relevance definition, which arises naturally  within the NRM framework.

\subsection{Simulated Attacker Subgraph in Social Network }
\label{sec:attacker_experiment}
\begin{figure}[t]
    \centering
    \includegraphics[width=0.8\linewidth]{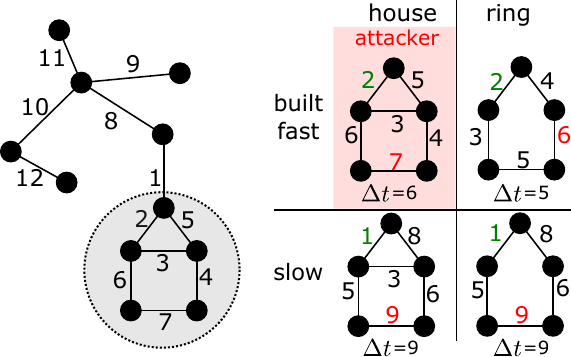}
    \caption{Class definitions in the  Attacker dataset. 
    The number next to each edge indicates the time at which it  was created.
Left: A temporal graph sample with an attacker motif marked with a circle. 
A graph containing at least one attacker is labeled positive.
Right:
A \textbf{house} subgraph built fast within \textbf{7 time steps} (i.e., $\Delta t
 ={\color{black} t_{max}}-{\color{black}t_{min}}+1\le 7$, where
 $t_{max}$ and $t_{min}$ are highlighted in red and green, respectively)
 is defined as an \textcolor{black}{attacker} subgraph, while the other three are defined as benign.  
}
    \label{fig:dataset_gen}
\end{figure}

Inspired by BA-2motif---a common synthetic benchmark dataset for evaluating XAI methods for static GNN \cite{schnake2020higher}---we synthesize a dataset of temporal graphs by adding 
temporal dependencies.
Specifically, we  generate a random Barabási–Albert graph attached with \emph{house} or \emph{ring} motif subgraph(s), and add a time-stamp to each edge (see \Cref{fig:dataset_gen} left).
We see each temporal graph as an artificial social network, where the nodes correspond to users and the event edges correspond to interactions between nodes, with or without \textit{attacker} subgraphs.
An attacker subgraph is a group of collaborative malicious users, and we assume that it consists of a fast-built dense sub-network, because the malicious users can easily interact with each other but not with genuine users.  
In our synthetic dataset, we label a house motif subgraph (consisting of \textit{6} edges) built within \textit{7} time steps as an attacker.
\Cref{fig:dataset_gen} left visualizes the definition of attacker and normal subgraphs.
We train an ETGNN for graph-level binary classification---whether a graph contains at least one attacker subgraph or not. 
For training the model, we generated 500 samples with half positive and half negative labels and randomly sampled 80\% for training and 20\% for testing. The test  accuracy is 99.0\%.

\begin{figure*}[t]
    \centering
    \includegraphics[width=\linewidth]{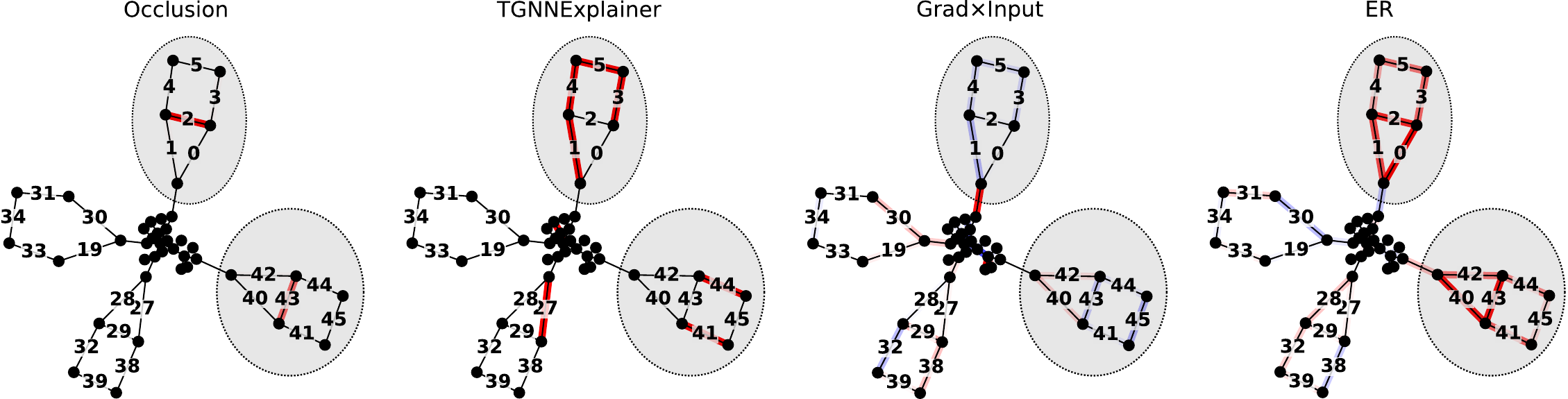}
    \caption{Explanation by baseline and our proposed methods.
    The ground-truth attacker subgraphs are marked with circles,
    and the numbers on the edges denote the timestamps.  Our ER successfully identifies both attackers, while the other methods fail.
    }
    \label{fig:attacker}
\end{figure*}

\subsubsection{Qualitative Visualization}
\label{sec:attacker_qualitative}

\Cref{fig:attacker} shows the relevance heatmaps obtained by 
the baseline methods and ER
for positive samples with two attacker subgraphs.
 Since  TGNNExplainer does not provide edge-level explanations, we visualize the most relevant 10-event subgraph in red with uniform intensity.
 The ground-truth attacker subgraphs are marked with circles.

We observe that 
our ER correctly identifies the attacker subgraph as positively relevant, while the other methods fail.
A possible reason of Occlusion's failure is that perturbing only one attacker should not change the prediction when another attacker exists. 
In contrast, 
ER identifies all attacker-related event edges as positively relevant, highlighting the effectiveness of our attribution approach. %
More results are provided in \ref{app:attacker_experiment}.

\subsubsection{Quantitative Evaluation}

We quantitatively evaluate the quality of the top-$k$ most relevant events detected by our methods and the baselines.  For TGNNExplainer, %
the top-$k$ relevant edges are defined as the edges of the most relevant subgraph consisting of $k$ edges.
The average scores over the top-$k$ events of four metrics,  Precision \eqref{eq:DefPrecision}, Recall \eqref{eq:DefRecall}, Pruning \eqref{eq:DefPerturb} and Activate \eqref{eq:DefActivate}, defined in \ref{sec:quant_eval_method},
are shown in \Cref{tab:experiment_results} (see \ref{app:qualitative_experiment} for the performance curves).
This evaluation was conducted on a test set of $450$  temporal graphs, where a half contain only one attacker subgraph, and the other half contain multiple attacker subgraphs.
We observe that ER significantly outperforms the other methods. Although Occlusion performs comparably well with ER on the graphs with only a single attacker, as shown in \ref{app:attacker_experiment},
it tends to fail to identify the relevant edges when multiple attackers exist, as discussed in \Cref{sec:attacker_qualitative}.

\subsection{Political Events Network}
\label{sec:icews_experiment}

ICEWS18 is a real-world event dataset that contains 468558 political events between various entities from 2018-01-01 to 2018-10-31 (one timestamp per day) \citep{jin-etal-2020-recurrent, DVN/28075_2015}. 
There are 304 time steps and on average 1541 events for each time step. Nodes have no feature, while events have 256 different tags (\textit{relation}s) indicating the event type.
For example,
with the notation introduced in \Cref{sec:TemporalGraph},
an event $e$: ``Kim Jong-Un$-$Threaten$\rightarrow$ United States on 2018-01-01'' is 
an interaction between the origin node $v_{\mathrm{o}}^{(e)}$: ``Kim Jong-Un'' and the destination node 
$v_{\mathrm{d}}^{(e)}$: ``United States'' happened at the time ``2018-01-01'' with the event type $\bftau$: ``Threaten''.

We consider the event-level multi-class classification task, where 
the type of event that happens  between two given entities is predicted. The dataset is one large temporal graph, of which the set of events are split into training and testing sets according to the time stamp of events: the events with the first 80\% time steps are used for training and the rest for testing. 
During training, we update the node memories with events up to time step $t$, then predict the events at  $t+1$ to calculate the loss function.
In the test time, we start from $t=0$, and update the node memories up to the final training event $t=T$, and make predictions for the test events.
Our model reaches about 33\% accuracy and 55\% Hit@3,\footnote{Hit@$k$ is defined as the percentage of test samples, for which the $k$ classes with highest predictive probability contains the target class.}
which is comparable to the model reported in Jin et al.~\cite{jin-etal-2020-recurrent}, although the authors used a different RE-Net architecture.

\subsubsection{Qualitative Results}

As a qualitative evaluation result, 
\Cref{tab:icews_vis} shows the most relevant events 
for the prediction of the event ``United Nations$-$Make an appeal or request$\rightarrow$ Legislature (Poland) on 2018-03-22.'' The table shows the 10 most relevant and the 3 least (or negatively most) relevant events detected by ER, Grad$\times$Input, and TGNNExplainer.  Occlusion is excluded due to its computational intractability, requiring one forward computation for each event.%
\footnote{
For example, to explain one event at $t=2000$ with Occlusion, the time needed is  $ \# \text{ of events} \times \text{forward computation time} \approx 1.6\times10^5 \times 30s \approx 57\text{days}$. 
}

Since no ground-truth explanation is available,  we chose the predicted events such that the relations with other events can be intuitively understood from news articles.
According to a news article,\footnote{\href{https://reproductiverights.org/over-200-sexual-reproductive-rights-ngos-call-for-polish-parliament-to-protect-womens-health-and-rights/}{https://reproductiverights.org/over-200-sexual-reproductive-rights-ngos-call-for-polish-parliament-to-protect-womens-health-and-rights/}}
the appeal or request was about stopping the bill that prohibits abortion. The explanation by our ER method successfully detects intuitively relevant events related to Poland's diplomatic cooperation with neighboring countries, for example, ``EU's accusation to Polish legislature'' and ``Polish citizen's demonstration or rally against Polish legislature.'' Besides, the result also highlighted several events that the UN is involved, which implies the way how the model understands UN's role. Our methods can not only detect temporally close events but also some events that occurred months before 
the predicted event. Compared to ours, Grad$\times$Input and TGNNExplainer provide less human-interpretable explanations, which are not very helpful for the users to understand the rationale of the model. We also observe that some events assigned negative relevance by Grad$\times$Input are, in fact, plausibly relevant for the prediction. This discrepancy may stem from the \textit{shattered gradient} effect
\citep{Montavon2019grad_vs_prop}, which introduces strong positive and negative fluctuations into the feature-wise relevance scores.
Additional examples of qualitative explanation are given in 
\ref{app:icews_exp}.

\newcolumntype{C}[1]{>{\centering\arraybackslash}p{#1}<{\rule[-0.5ex]{0pt}{3ex}}}
\newcolumntype{L}[1]{>{\arraybackslash}p{#1}<{\rule[-0.5ex]{0pt}{3ex}}}
\newcolumntype{R}[1]{>{\raggedleft\arraybackslash}p{#1}<{\rule[-0.5ex]{0pt}{3ex}}}
\begin{table*}[t]
    \tiny
    \centering
    \begin{tabular}{R{.2\textwidth}cL{.33\textwidth}cC{.18\textwidth}|l}
    \textbf{Method: ER} \\
    \hline
    Event &&&&& \#D ago\\
    \hline
\rowcolor{Salmon!100} {Legislature (Poland)} & {--}& {Engage in diplomatic cooperation} & {$\rightarrow$}& {Lithuania}  & {11}\\
\rowcolor{Salmon!100} {Lithuania} & {--}& {Engage in diplomatic cooperation} & {$\rightarrow$}& {Legislature (Poland)}  & {11}\\
\rowcolor{Salmon!73} {United Nations} & {--}& {Appeal for change in leadership} & {$\rightarrow$}& {Joseph Kabila}  & {80}\\
\rowcolor{Salmon!41} \textbf{European Commission} & \textbf{--} & \textbf{Accuse} & \textbf{$\rightarrow$} & \textbf{Legislature (Poland)} & \textbf{71} \\
\rowcolor{Salmon!38} \textbf{Citizen (Poland)} & \textbf{--} & \textbf{Demonstrate or rally} & \textbf{$\rightarrow$} & \textbf{Legislature (Poland)} & \textbf{67} \\
\rowcolor{Salmon!37} \textbf{Legislature (Poland)} & \textbf{--} & \textbf{Investigate} & \textbf{$\rightarrow$} & \textbf{Citizen (Poland)} & \textbf{68} \\
\rowcolor{Salmon!27} David Granger &--& Make statement &$\rightarrow$& United Nations  & 80\\
\rowcolor{Salmon!27} United Nations &--& Make statement &$\rightarrow$& Military (Myanmar)  & 80\\
\rowcolor{Salmon!27} \textbf{Citizen (Poland)} & \textbf{--} & \textbf{Demonstrate or rally} & \textbf{$\rightarrow$} & \textbf{Legislature (Poland)} & \textbf{69} \\
\rowcolor{Salmon!24} United Nations &--& Make an appeal or request &$\rightarrow$& Actor (Congo)  & 80\\
$\cdots$\\
\rowcolor{cyan!1} Lithuania &--& Consult &$\rightarrow$& Estonia  & 1\\
\rowcolor{cyan!1} Latvia &--& Consult &$\rightarrow$& Lithuania  & 1\\
\rowcolor{cyan!1} Lithuania &--& Consult &$\rightarrow$& Latvia  & 1\\
\bottomrule
\\
\textbf{Method:  Grad$\times$Input}\\
\hline
Event &&&&& \#D ago\\
\hline
\rowcolor{Salmon!100} United Nations &--& Appeal for change in leadership &$\rightarrow$& Joseph Kabila  & 80\\
\rowcolor{Salmon!63} Andrzej Duda &--& Make statement &$\rightarrow$& Poland  & 80\\
\rowcolor{Salmon!56} \textbf{Citizen (Poland)} &--& \textbf{Demonstrate or rally} &$\rightarrow$& \textbf{Legislature (Poland)}  & 67\\
\rowcolor{Salmon!44} \textbf{United States} &--& \textbf{Make statement} &$\rightarrow$& \textbf{Legislature (Poland)  }& 50\\
\rowcolor{Salmon!42} Saulius Skvernelis &--& Make an appeal or request &$\rightarrow$& Citizen (Poland)  & 76\\
\rowcolor{Salmon!39} Saulius Skvernelis &--& Express intent to meet or negotiate &$\rightarrow$& Citizen (Poland)  & 76\\
\rowcolor{Salmon!37} Lawmaker (Russia) &--& Consider policy option &$\rightarrow$& Russia  & 80\\
\rowcolor{Salmon!32} Czech Republic &--& Engage in diplomatic cooperation &$\rightarrow$& Russia  & 80\\
\rowcolor{Salmon!32} Russia &--& Engage in diplomatic cooperation &$\rightarrow$& Czech Republic  & 80\\
\rowcolor{Salmon!28} United Nations &--& Make an appeal or request &$\rightarrow$& Actor (Congo)  & 80\\
$\cdots$\\
\rowcolor{cyan!51} \textbf{Citizen (Poland)} &--& \textbf{Impose embargo, boycott, or sanctions} &$\rightarrow$& \textbf{Poland}  & 79\\
\rowcolor{cyan!52} Member of Legislative  (Lithuania) &--& Make an appeal or request &$\rightarrow$& Legislature (Poland)  & 21\\
\rowcolor{cyan!82} \textbf{Ethnic Group (Poland)} &--& \textbf{Make an appeal or request }&$\rightarrow$& \textbf{Legislature (Poland)  }& 45\\
    \bottomrule
\\
\textbf{Method:  TGNNExplainer}\\
\hline
Event &&&&& \#D ago\\
\hline
\rowcolor{Salmon!50} United Nations &--& Praise or endorse &$\rightarrow$& International Committee of the Red Cross  & 2\\
\rowcolor{Salmon!50} United Nations &--& Investigate &$\rightarrow$& Government (Sweden)  & 1\\
\rowcolor{Salmon!50} Foreign Affairs (Turkey) &--& Accuse &$\rightarrow$& United Nations  & 2\\
\rowcolor{Salmon!50} United States &--& Provide economic aid &$\rightarrow$& United Nations  & 2\\
\rowcolor{Salmon!50} United Nations &--& Criticize or denounce &$\rightarrow$& Foreign Affairs (Turkey)  & 1\\
\rowcolor{Salmon!50} Revolutionary Armed Forces of Colombia &--& Provide military aid &$\rightarrow$& United Nations  & 2\\
\rowcolor{Salmon!50} United Nations &--& Make statement &$\rightarrow$& Pierre Nkurunziza  & 2\\
\rowcolor{Salmon!50} United Nations &--& Consult &$\rightarrow$& Abd al-Rab Mansur al-Hadi  & 2\\
\rowcolor{Salmon!50} Abd al-Rab Mansur al-Hadi &--& Consult &$\rightarrow$& United Nations  & 2\\
\rowcolor{Salmon!50} United States &--& Praise or endorse &$\rightarrow$& United Nations  & 2\\
    \bottomrule
    \end{tabular}
    \vspace{.3cm}
    \caption{The 10 most positively relevant and the 3 least (or most negatively) relevant events detected by our ER,  Grad$\times$Input, and TGNNExplainer for  predicting the event ``United Nations$-$Make an appeal or request$\rightarrow$ Legislature (Poland)'' 
   (TGNNExplainer does not provide the least relevant events).
    Red and blue color with different intensity indicates the positive and negative relevance, respectively. ER does not attribute significantly negative relevance and therefore the least relevance is neutral (white). '$\#D$ ago' indicates how many days before the predicted event the corresponding event occurred. We marked in bold the  events that are assumed to be 
    relevant.}
    \label{tab:icews_vis}
\end{table*}

\subsubsection{Quantitative Evaluation}

We preformed pruning and activation tests for the top-20 most relevant events on 50 randomly sampled events from the ones correctly predicted.
Average scores over top-20 are shown in \Cref{tab:experiment_results} (the performance curves can be found in
\ref{app:qualitative_experiment}). 
Our ER methods significantly outperform the baselines in the  pruning test, while slightly worse than Grad$\times$Input in the activation test.

 {
\subsection{Computational Time}

Here, we report on computation time of our method and the baselines. We run the experiments on an M1Pro CPU.

For a given temporal graph,
ER performs a forward computation to prepare the propagation matrices and the output relevance, and then a backward computation for the relevance propagation. We denote the computation time for forward and backward computation as $t_{\rightarrow}^{\mathrm{ER}}$ and $t_{\leftarrow}^{\mathrm{ER}}$, respectively. The backward computation can be costly for large temporal graphs. For example, for the temporal graph from ICEWS18 with 468558 events, $t_{\rightarrow}^{\mathrm{ER}} \approx 2 \text{ min}$ and $t_{\leftarrow}^{\mathrm{ER}} \approx 18 \text{ min}$. However, for small graphs in the Attacker dataset, $t_{\rightarrow}^{\mathrm{ER}} \approx 0.003 \text{ sec}$ and $t_{\leftarrow}^{\mathrm{ER}} \approx 3.5 \text{ sec}$. 

 Occlusion performs a forward computation for each occlusion pattern.  This is costly when we have to compute the relevance of all events to  identify the most relevant ones in a temporal graph.
For the ICEWS18 dataset,
a single forward computation 
takes $\sim 1\text{ min}$, and thus the total computation time is $468558\times 1 \text{ min}\approx 325\text{ day}$, which is infeasible. 
For small graphs 
with only $\sim 50$ events in the Attacker dataset, Occlusion takes only $\sim 1\text{ sec}$.

Grad$\times$Input requires only a single forward pass and backward relevance propagation, which 
are generally inexpensive with automatic differentiation libraries. It takes $\sim 5 \text{ min}$ and $\sim 0.03\text{ sec}$ for the ICEWS18 and Attacker datasets, respectively.

TGNNExplainer requires training an MLP as a `navigator' and then running a Monte-Carlo Tree Search. For the latter part, the most time consuming process is in the reward calculation, which is equal to a single forward computation of the model on a subgraph. This process is performed once for each rollout. By default, the rollout times is set to 500  and the initial event subgraph size is 25 as in Xia et al.~\cite{DBLP:conf/iclr/XiaLS0D0023}, which is by orders of magnitude smaller than the original temporal graph. Therefore the whole process can be finished in a few seconds. 

\Cref{tab:runtime_comparison} summarizes the computation time.

\begin{table}[htbp]
    \centering
    \begin{tabular}{l|rr}
        Method & ICEWS18 & Attacker \\
        \hline
        ER (ours) & 20 min & 3.5 sec\\
        Occlusion & infeasible & 1 sec \\
        Grad$\times$Input & 5 min & 0.03 sec\\
        TGNNExplainer & 10 sec & 3 sec\\
    \end{tabular}
    \caption{Approximate runtime for our method and baselines on a big graph (ICEWS18) and small graphs (Attacker). ER, providing the best explanation in our qualitative and quantitative experiments, requires a reasonable amount of computation time. For TGNNExplainer we only consider the MCTS searching process.}
    \label{tab:runtime_comparison}
\end{table}
}

\section{Conclusion}
\label{sec:Conclusion}

Time-series data play a central role in many real-world applications, ranging from scientific monitoring to event prediction in complex systems. While modeling such data with deep neural networks has achieved strong predictive performance, their internal decision mechanisms often remain opaque. In particular, explaining temporal graph neural networks is challenging due to their dependence on evolving latent representations and long-range event interactions, which renders existing explainability techniques incomplete or insufficiently faithful.

In this paper, we focused on explainability for Event-based Temporal Graph Neural Networks (ETGNNs), and proposed a novel event relevance (ER) definition built upon the Normalized Relevance Measure (NRM) framework. Unlike existing methods that overlook latent memory evolution, our approach accounts for the full event-induced information flow, enabling more faithful  explanations. To make NRM applicable to such complex architectures, we introduced a modularized relevance decomposition scheme that supports hierarchical relevance definition. Furthermore, our method naturally extends to higher-order attribution, allowing the joint relevance of multiple events to be quantified and thereby capturing interactions among events for more fine-grained explanations.

Our experiments on artificial and real-world datasets showed that our proposed approach achieves state-of-the-art performance, producing both qualitatively interpretable heatmaps and quantitatively faithful explanations.
In addition, the joint relevance analysis successfully captures long-range interactions, which are essential for accurate modeling in tasks such as infection-chain prediction. 

In future work, we plan to develop a tool that further facilitates the construction of relevance structures based on the proposed modular decomposition, ultimately enabling partial automation from the specification of forward computations and LRP rules. Further improvements toward more efficient implementations, for example, via the forward-hook trick \cite{schnake2020higher,DBLP:journals/pieee/SamekMLAM21}, are also of interest. These techniques would further extend the applicability of the NRM framework to a broader class of models.

\section*{CRediT authorship contribution statement}
Ping Xiong: Writing – review \& editing, Writing – original draft,
Visualization, Validation, Software, Methodology, Investigation, Formal analysis, Conceptualization; 
Thomas Schnake: Writing – review \& editing, Writing – original draft, Visualization, Methodology, Investigation, Formal analysis, Conceptualization; 
Klaus-Robert M\"{u}ller: Writing – review \& editing, Supervision, Resources, Methodology, Funding acquisition, Conceptualization; 
Shinichi Nakajima:
Writing – review \& editing, Writing – original draft, Supervision, Project
administration, Methodology, Investigation, Formal analysis, Conceptualization.

\section*{Declaration of competing interest}
The authors declare that they have no known competing financial
interests or personal relationships that could have appeared to influence
the work reported in this paper.

\section*{Acknowledgments}
The authors thank Grégoire Montavon for helpful discussions and valuable suggestions during this research.
This work was funded by the German Ministry for Education and Research as BIFOLD - Berlin Institute for the Foundations of Learning and Data (ref. BIFOLD25B). 
Klaus-Robert M\"{u}ller was partly supported by the Institute of Information \& Communications Technology Planning \& Evaluation (IITP) grant funded by the Korea government (MSIT) (No. RS-2019-II190079, Artificial Intelligence Graduate School Program, Korea University) and grant funded by the Korea government (MSIT, No. RS-2024-00457882, AI Research Hub Project), and Hector Fellow Academy. Furthermore this work is partly supported by DFG and by the Korea University Grant. Thomas Schnake is a postdoctoral fellow at the University of Toronto in the Eric and Wendy Schmidt AI in Science Postdoctoral Fellowship Program, a program of Schmidt Sciences.

\bibliographystyle{elsarticle-num}
\bibliography{main.bib}

\appendix
\setcounter{figure}{0}
\renewcommand{\thetheorem}{\Alph{section}.\arabic{theorem}}

\allowdisplaybreaks

\section{Modularized NRM Definition for LSTM}
\label{app:lstm}

\begin{figure*}
    \centering
    \includegraphics[width=0.73\linewidth]{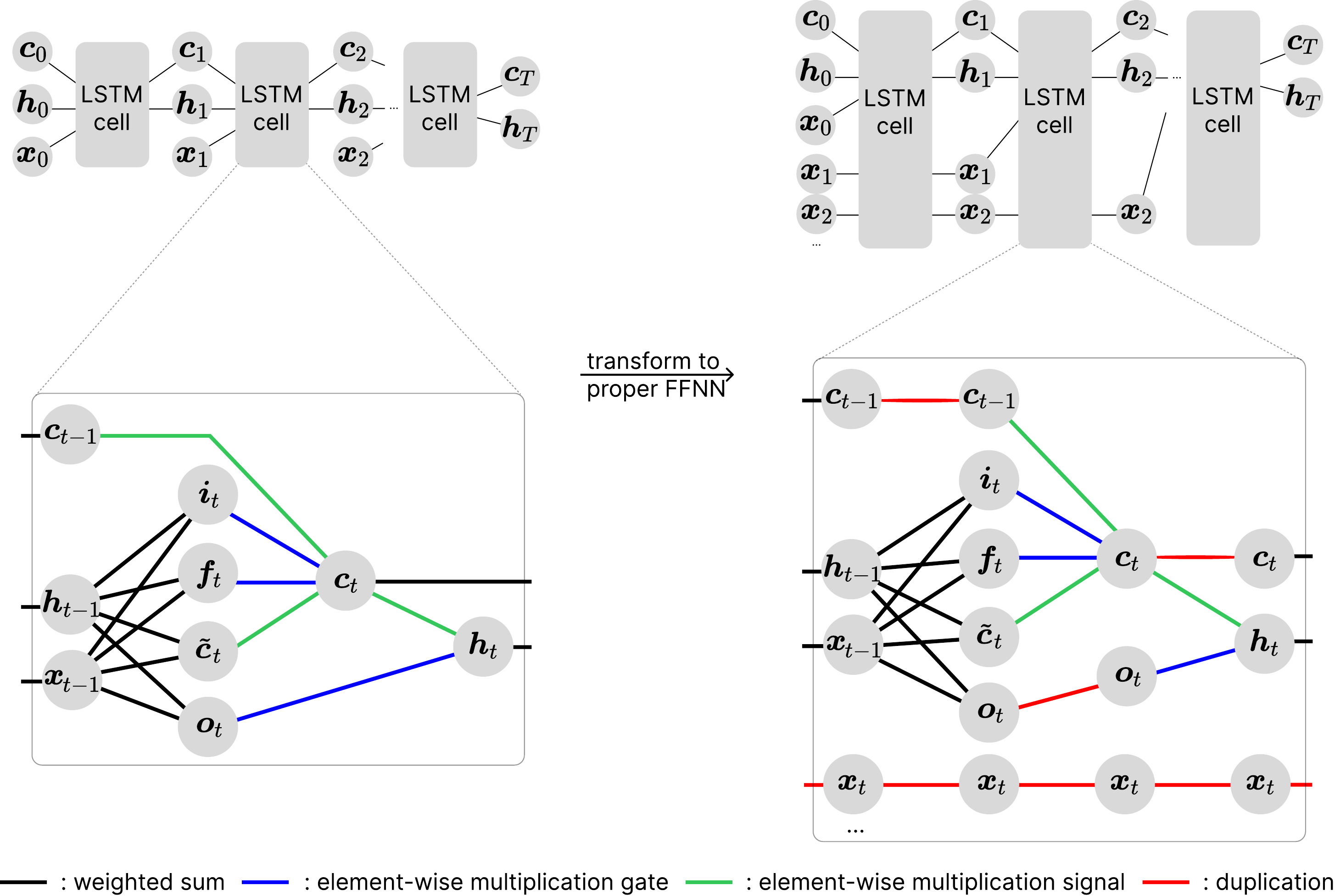}
    \caption{Application of NRM to LSTM with modularization. Left: original LSTM with skip connections and inputs in intermediate layers. Right: LSTM transformed into a proper FFNN.
    LSTM can be seen as a series connection of LSTM cell modules.
    The color of connections indicates the type of input-output relation.
    }
    \label{fig:app_lstm}
\end{figure*}

Here we give details of example application of the NRM framework to an LSTM network (\Cref{fig:app_lstm} left).
Following Xiong et al.~\cite{NRMpaper}, we first convert the network into a proper FFNN by unfolding the recurrent architecture, copying intermediate inputs and outputs to the input and output layers, respectively, and eliminating the skip connections by duplicating neurons across the skipped intermediate layers (\Cref{fig:app_lstm} right). 

Viewing each LSTM cell as a module, we can decompose the relevance of the entire LSTM network as a $T$-layer modular-wise network:
\begin{align}
    R_{\mathrm{LSTM}}(\bfn^{(0: T-1)} | n^{(T)}) = \prod_{t=1}^{T}R_{\mathrm{LSTM-Cell}}(n^{(t-1)}| n^{(t)}),\label{eq:Relevance.LSTMNetwork}
\end{align}
where $n^{(\cdot)}$ specifies a neuron in each layer as
\begin{align}
n^{(t)} \in \mcS^{(t)}( \{ \bfh_{t}, \bfc_{t}, \{\bfx_{t'}\}_{t'=t}^{T-1} \}).
\notag
\end{align}
Here, 
$\mcS^{(l)}({{\mcX}})$ denotes the set of neurons in the $l$-th layer that represent a set $\mcX$ of the variables,
and we used the hierarchical layer indexing convention---i.e., layer indices are defined locally in each hierarchy and each module for convenience---as explained in \Cref{sec:Modularization.MOdularization}.

According to the layered structure within each LSTM cell module (see the right bottom plot in \Cref{fig:app_lstm}),
the relevance 
of an LSTM-cell is decomposed as a $3$-layer network as
\begin{align}
 &   R_{\mathrm{LSTM-Cell}}(n^{(0)}, n^{(1)}, n^{(2)}| n^{(3)}) = 
    \notag\\
&    \qquad
    R_{\mathrm{Step1}}(n^{(0)}| n^{(1)})
    R_{\mathrm{Step2}}(n^{(1)}| n^{(2)})
    R_{\mathrm{Step3}}(n^{(2}| n^{(3)}),
    \label{eq:LSTMCellDecomposition}
\end{align}
where $n(\cdot)$ specifies a neuron in each layer as
\begin{align}
    n^{(0)} &\in \mcS^{(0)}( \{ \bfh_{t-1}, \bfc_{t-1}, \{\bfx_{t'}\}_{t'=t-1}^{T-1} \}),\notag\\
    n^{(1)} &\in \mcS^{(1)}( \{ \bfc_{t}, \bfi_{t},\bff_{t}, \tilde{\bfc}_{t},\bfo_{t},\{\bfx_{t'}\}_{t'=t}^{T-1} \}),\notag\\
    n^{(2)} &\in \mcS^{(2)}( \{ \bfc_{t},\bfo_{t},\{\bfx_{t'}\}_{t'=t}^{T-1} \}),\notag\\
    n^{(3)} &\in \mcS^{(3)}( \{ \bfc_{t},\bfh_{t},\{\bfx_{t'}\}_{t'=t}^{T-1} \}).\notag
\end{align}

Each factor in Eq.\eqref{eq:LSTMCellDecomposition} is defined, following the general NRM construction (see Section 3.3 in Xiong et al.~\cite{NRMpaper}).
Let us denote the variables in the $(l-1)$-th layer as $\{\bfmu^{(l-1, j)} \}_{j=1}^J$, and the variables in the $l$-th layer as $\{\bfnu^{(l, k)}  \}_{k=1}^{K}$ with the forward process in the following form:
 \begin{align}
\bfnu^{(l, k)} =    \bfg^{(k)} (\{ \bfW^{(l, j \to k)\T} \bfmu^{(l-1, j)} \}_{j=1}^J) 
\label{eq:ComplicatedForwardProcessGeneral}
 \end{align}
 for $k = 1, \ldots, K$.
Here $\bfW^{(l, j \to k)} \in \mathbb{R}^{N^{(l-1, j)}\times N^{(l, k)} }$ is the network weight parameter matrix responsible for the conversion from $\bfmu^{(l-1, j)}$ to $\bfnu^{(l, k)}$.
For example, for the first layer of the LSTM cell (see the bottom right of \Cref{fig:app_lstm}),
the input variables are
$\{\bfmu^{(l-1, j)} \}_{j=1}^4
= \{\bfc_{t-1}, \bfh_{t-1}, \bfx_{t- 1}, \bfx_t\}$, the output variables are $\{\bfnu^{(l, k)}  \}_{k=1}^{6} = \{\bfc_{t-1}, \bfi_t, \bff_{t}, \tilde{\bfc}_{t}, \bfo_t, \bfx_t\}$,
and the weight parameter matrices correspond to the identity matrix $\bfI$ for the duplication links, while they correspond to the conversion matrices, $\bfW_{\cdot}$ and $\bfU_{\cdot}$ in Eqs.\eqref{eq:LSTM.F.first}--\eqref{eq:LSTM.F.forth}, for other links.

To define the consecutive conditional relevance between the $(l-1)$-th and the $l$-th layers, we first define unnormalized relevance propagation matrices $\{\widetilde{\bfT}^{(l, j \leftarrow k)}\}$ between each pair of variables.  For the duplication links (depictd in red in \Cref{fig:app_lstm}), we simply set the propagation matrices to 
\begin{align}
 \widetilde{\bfT}^{(l, j \leftarrow k)} = 
 \bfI.
\notag
\end{align}
For the other links (depicted in other colors), the matrices depend on the choice of the LRP rule.
For example, if we choose the LRP-$\gamma$ rule, we set
\begin{align}
 \widetilde{T}_{n, n'}^{(l, j \leftarrow k)} = 
 \mu^{(l-1, j)}_{n} W^{(l, j \to k)\uparrow}_{n, n'} + \varepsilon,
\label{eq:UnnormalizedPropagationMatrix}
\end{align}
where
$ \bfW ^{(l)\uparrow} \in \mathbb{R}^{\numneu{l-1} \times \numneu{l}}$ is a modified network weight parameter at the $l$-th layer such that
$\bfW ^{\uparrow} := \bfW  + \gamma\cdot \max(0, \bfW  )$
with the maximization operator applying entry-wise, and  $\varepsilon > 0$ is a small constant for stabilization.

Then, the complete unnormalized propagation matrix $\widetilde{\bfT}^{(l)} $ is  defined as
\begin{align}
    \widetilde{\bfT}^{(l)}
    & = 
    \begin{pmatrix}
      \alpha^{(l, 1 \leftarrow 1)}  \widetilde{\bfT}^{(l, 1 \leftarrow 1)} & \cdots &    \alpha^{(l, 1 \leftarrow k )}  \widetilde{\bfT}^{(l, 1 \leftarrow K)} \\
        \vdots &\ddots  & \vdots \\
     \alpha^{(l, J \leftarrow 1)}   \widetilde{\bfT}^{(l, J \leftarrow 1)} & \cdots &  \alpha^{(l, J \leftarrow K)}\widetilde{\bfT}^{(l, J \leftarrow K)}     \end{pmatrix},
        \label{eq:WholeUnnormalizedPropagationMatrix}
\end{align}
where $\{  \alpha^{(l,  j \leftarrow k )} \in \mathbb{R}\} $ are controllable coefficients specifying the relevance distribution from the output to the input.
Those coefficients are often set to $\alpha^{(l, j \leftarrow k)}  = 1$ for the weighted sum links (black) and the element-wise multiplication signal links (green), while they are set to $\alpha^{(l, j \leftarrow k)}  = 0$ for the  element-wise multiplication gate links (blue), which corresponds to the LRP-all rule \citep{arras2017recurrent}.

With the whole matrix \eqref{eq:WholeUnnormalizedPropagationMatrix} defined, 
the consecutive conditional relevance is given by the column-wise normalization, i.e.,
\begin{align}
R(n^{(l-1)} | n^{(l)})
 &  = \textstyle
  \frac{   \widetilde{T}^{(l)}_{n^{(l-1)}, n^{(l)}}  }{  \sum_{n'^{(l-1)}}    \widetilde{T}^{(l)}_{n'^{(l-1)}, n^{(l)}}   } .
 \label{eq:ConsecutiveConditionalRelevanceContruction}
 \end{align}
 Applying this procedure to all three layers of the LSTM cell gives the three factors in 
the right-hand side of Eq.\eqref{eq:LSTMCellDecomposition}.

The output relevance of the LSTM network is defined as follows.
We first set an
\emph{unnormalized output relevance vector} based on the entire network output $\bff\left( \{\bfx_t\}_{t=0}^{(T-1)}\right)$; for example,  
\begin{align}    
\widetilde{\bfr}^{(T)} = \bff\left( \{\bfx_t\}_{t=0}^{(T-1)}\right) ,
\label{eq:OutputRelevanceExample}
\end{align}
and then normalize it:
 \begin{align}
 R(n^{(T)})
  & = \textstyle
  \frac{
\widetilde{r}^{(T)}_{n^{(T)}}
  }
  {
   \sum_{n'^{(T)}}
\widetilde{r}^{(T)}_{n'^{(T)}}
  }.
 \label{eq:OutputRelevanceContruction}
 \end{align}

Eqs.\eqref{eq:Relevance.LSTMNetwork} and \eqref{eq:OutputRelevanceContruction}, which define the top-level module-wise relevance structure, together with  Eqs.\eqref{eq:LSTMCellDecomposition} and \eqref{eq:ConsecutiveConditionalRelevanceContruction},
which define the relevance structure within each LSTM cell, specify all full joint (or walk) relevances of the LSTM network, thereby defining the relevance of any set of neurons.

\section{Example of Parallel Modularization}
\label{sec:A.ExampleModularization}

\begin{figure}[t]
    \centering
   \includegraphics[width=0.8\linewidth]{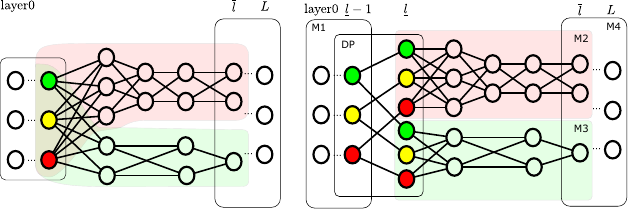}
   \caption{An example network consisting of four modules.  Left: the parallel modules (red and green shadows) share the same input neurons (colored).  Right: DuPlication (DP) layer is inserted such that the parallel modules have no overlapping, and thus the top-level relevance is decomposed as Eq.\eqref{eq:A.ParallelMessagePassingExampleDecomposition}.
   }
    \label{fig:HierarchicalLayers}
\end{figure}

Consider the network shown in \Cref{fig:HierarchicalLayers} (left),
which
consists of four modules connected in series and in parallel.
We first make the two parallel modules, i.e., Module 2 (red shadow) and Module 3 (green shadow), non-overlapping by inserting a DuPlication (DP) layer, as shown in \Cref{fig:HierarchicalLayers} (right). 
Then, following the series \eqref{eq:SeriesDecompositionConditional} and parallel \eqref{eq:ParallelDecompositionConditional} connection rules and hierarchical layer indexing, the top-level relevance is decomposed as 
\begin{align}
R(n^{(0)}, n^{(1)}, n^{(2)}, n^{3)} | n^{(4)})
&= 
R_{\mathrm{M1}}(n^{(0)} | n^{(1)}) 
R_{\mathrm{DP}}(n^{(1)}|n^{(2)}) 
 \notag\\
& \hspace{-26mm}
\left\{
R_{\mathrm{M2}}^{ \subseteq \mcS^{(2:3)}}
(n^{(2)} | n^{(3)} )
+
R_{\mathrm{M3}}^{ \subseteq \overline{\mcS}^{(2:3)}}
(n^{(2)}| n^{(3)} ) 
\right\}
R_{\mathrm{M4}}(n^{(3)} | n^{(4)}),
\label{eq:A.ParallelMessagePassingExampleDecomposition}
\end{align}
where each factor on the right-hand side can be defined locally using the layer indices of the corresponding module.
Note that the layer indices need not be synchronized across parallel modules, since the  relevance is normalized within each module.  Consequently, the sum of the relevance at an arbitrary layer in M2 and that at an arbitrary layer in M3 is always equal to one.

Given a target quantity, for example, the input feature relevance $R(n^{(0)} | n^{(4)})$ with respect to an output,
the message passing should be first applied to the last factor $ R_{\mathrm{M4}}(n^{(3)} | n^{(4)})$, propagating the relevance from its output to the input.  Then, the message passing for the local relevances 
$R_{\mathrm{M2}}^{ \subseteq \mcS^{(2:3)}}
(n^{(2)} | n^{(3)} )$
and $R_{\mathrm{M3}}^{ \subseteq \overline{\mcS}^{(2:3)}}
(n^{(2)}| n^{(3)} ) $ should be applied
in parallel,
 giving the 
 conditional relevance $R(n^{(2)}| n^{(4)} )$.
Finally, the message passing for $R_{\mathrm{DP}}(n^{(1)}|n^{(2)}) $ and 
$R_{\mathrm{M1}}(n^{(0)} | n^{(1)}) $ is applied sequentially, giving the target quantity $R(n^{(0)} | n^{(4)}) $.

\section{Details of Relevance Definition for ETGNNs}
\label{sec:A.ETGNN.JointRelevanceDefinition}

In this section, we specify each factor in Eq.\eqref{eq:GlobalDecomposition} to complete the relevance definition of an entire ETGNN.
For simplicity, we describe the relevance of a simple ETGNN consisting of the identity message function, the sum aggregation function, the GRU node memory updating function, the identity Emb module and a linear decoder for node level prediction. We further assume that the batch size is one, so that each batch corresponds to a single event, and that no node-level event exists. 
One can define the relevance of  general ETGNNs in the same way with more steps and hierarchy.

\subsection{Event Processing (EP) Module}
\label{sec:ep_module}

\begin{figure}[!t]
    \centering
    \includegraphics[width=\linewidth]{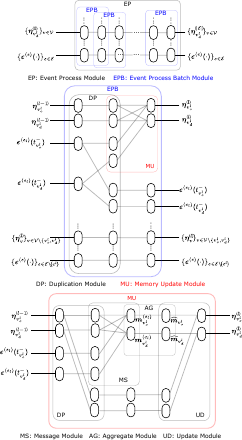}
    \caption{Hierarchical module structure of EP module (top), EPB module (middle), and MU module (bottom).}
    \label{fig:EventProcessingModuleSimple}
\end{figure}

The simplified EP module can be seen as a $|\mathcal E|$-layer NN, where the node memories $\{\bfeta_v^{(l)}\}$ evolve through interactions induced by a single event $e_l$ at each layer. Specifically,  in the $l$-th layer, the node memories of the origin node $v_{\mathrm{o}}^{l} := v_{\mathrm{o}}^{(e_l)}$ and the destination node $v_{\mathrm{d}}^{l} := v_{\mathrm{d}}^{(e_l)}$ are updated as follows:
\begin{align}
    \bfeta_v^{(l)} &=
    \begin{cases}
        \bfeta_v^{(l-1)},  &v\notin \{v_{\mathrm{o}}^l, v_{\mathrm{d}}^l\},\\
        \text{GRU}(\bfm^{(e_l)}_v, \bfeta_{v}^{(l-1)}),& v\in \{v_{\mathrm{o}}^l, v_{\mathrm{d}}^l\},\\
    \end{cases}
    \end{align}
    where
    \begin{align}
    \bfm^{(e_l)}_{v_{\mathrm{o}}^l} &= \left[\bfeta_{v_{\mathrm{o}}^l}^{(l-1)} || \bfeta_{v_{\mathrm{d}}^l}^{(l-1)} || \bfepsilon^{(e_l)}(t_{v_{\mathrm{o}}^l}^-) \right], \\
    \bfm^{(e_l)}_{v_{\mathrm{d}}^l} &= \left[\bfeta_{v_{\mathrm{d}}^l}^{(l-1)} || \bfeta_{v_{\mathrm{o}}^l}^{(l-1)} || \bfepsilon^{(e_l)}(t_{v_{\mathrm{d}}^l}^-) \right],
\end{align}
are the two \emph{messages} induced by the event $e_l$.
Here $[\cdot||\cdot]$ denotes a concatenating operation. 
In a proper FFNN representation (see \Cref{fig:EventProcessingModuleSimple} top),
event embeddings $\bfepsilon^{(e_l)} (\cdot)$ are additionally copied from the first to the last layer.

For the top-level relevance of the EP module, corresponding to the first factor in Eq.\eqref{eq:GlobalDecomposition},
\begin{align}
R_{\mathrm{EP}}(n^{(0)} | n^{(1)} ),
\notag
\end{align}
the neuron sets of input and output variables are
\begin{align}
  n^{(0)}  \in  \mathcal S^{\text{in}}_{\text{EP}} &= \mcS(\{ \bfeta_v^{(0)} | v\in\mathcal V\} \cup  \underbrace{\{ \bfepsilon^{(e_l)}(t_{v}^-)\}_{v\in\{v_{\mathrm{o}}^l,v_{\mathrm{d}}^l\}, l = 1,\dots,|\mathcal E|}}_{\vec{\mathcal E}}),\notag \\
    n^{(1)}  \in   \mathcal S^{\text{out}}_{\text{EP}} &= \mcS\left(\{ \bfeta_v^{(|\mathcal E|)} \}_{v\in\mathcal V} \cup  \vec{\mathcal E}\right).\notag 
\end{align}

\subsection{Event Processing Batch (EPB) Module}

The EP module consists of a series of EP Batch (EPB) modules (\Cref{fig:EventProcessingModuleSimple} top).  Accordingly, its relevance is decomposed as
\begin{align}
    R_{\text{EP}}(n^{(0)},\dots,n^{(|\mathcal E|-1)}|n^{(|\mathcal E|)}) = \prod_{l=1}^{|\mathcal E|} R_{\text{EPB}}(n^{(l-1)}|n^{(l)}),
\end{align}
where the neuron sets in each layer is 
\begin{align}
    n^{(l)}
    \in
    \mathcal S^{\text{out}}_{\text{EPB},l} &= \mcS\left(\{ \bfeta_v^{(l)} \}_{v\in\mathcal V} \cup  \vec{\mathcal E}\right)\notag 
\end{align}
(see \Cref{fig:EventProcessingModuleSimple} middle).

The relevance of the EPB module can be factorized as
\begin{align}
&R_{\mathrm{EPB}}( n^{(0)}, n^{(1)}|n^{(2)})
=
R_{\mathrm{DP}}( n^{(0)}|n^{(1)}) \cdot \notag \\
& \qquad \left(
R_{\mathrm{MU}}^{ \subseteq \mathrm{MU}}( n_{\mathrm{MU}}^{(1)}  | 
n_{\mathrm{MU}}^{(2)})
+ R_{\mathrm{ID}}^{ \subseteq \overline{\mathrm{MU}}}( n_{\overline{\mathrm{MU}}}^{(1)}  | n_{\overline{\mathrm{MU}}}^{(2)})
\right),
\label{eq:EPAdditiveDecomposition1}
\end{align}
where $n_{\mbox{...}}^{(\cdot)}$ specifies a neuron in each local layer as
\begin{align}
n^{(0)} &\in \mathcal S^{\text{in}}_{\text{EPB},l}, \notag \\
n^{(1)}
&\in
\mcS\left(
MU
\cup
\overline{MU}
\right),\notag \\
MU &= \{\bfeta_{v_{\mathrm{o}}^l}^{(l-1)}, \bfeta_{v_{\mathrm{d}}^l}^{(l-1)}, \bfepsilon^{(e_l)}(t^-_{v_{\mathrm{o}}^l}), \bfepsilon^{(e_l)}(t^-_{v_{\mathrm{d}}^l})\},\notag 
\\
\overline{MU} &= \vec{\mcE}\cup \{ \bfeta_{v^l}^{(l-1)}\}_{v\in\mcV\backslash \{v_{\mathrm{o}}^l, v_{\mathrm{d}}^l \}},
\notag\\
n^{(1)}_{MU} &\in \mcS(MU), n^{(1)}_{\overline{MU}} \in \mcS(\overline{MU}),\label{eq:EPAdditiveDecompositionNeurons} \\
n^{(2)}
&\in
\mcS\left(
MU'
\cup
\overline{MU'}
\right), 
\notag \\
MU' &= \{\bfeta_{v_{\mathrm{o}}^l}^{(l)}, \bfeta_{v_{\mathrm{d}}^l}^{(l)}\},\notag 
\\
\overline{MU'} &= \vec{\mcE}\cup \{ \bfeta_{v^l}^{(l)}\}_{v\in\mcV\backslash \{v_{\mathrm{o}}^l, v_{\mathrm{d}}^l \}},
\notag\\
n^{(2)}_{MU} &\in \mcS(MU'), n^{(2)}_{\overline{MU}} \in \mcS(\overline{MU'}).\notag
\end{align}
Here, $R_{\mathrm{MU}}^{ \subseteq \mathrm{MU}}(\cdots)$
is the local relevance \eqref{eq:LocalRelevance}
 of the MU module,
 which is defined below.
$R_{\mathrm{DP}}(\cdots)$ is the relevance of a DP module, which \emph{parallelizes} the MU and its complement,
and  $R_{\mathrm{ID}}^{ \subseteq \overline{\mathrm{MU}}} (\cdots)$
is the local relevance of an IDentical copying process.

\subsection{Memory Update (MU) Module}
\label{sec:MUModule}

An MU module consists of DP, MS, AG and UD sub-modules as shown in \Cref{fig:EventProcessingModuleSimple} (bottom). The relevance of the MU module is defined as follows:
\begin{align}
    & R_{\mathrm{MU}}(n^{(0)}, n^{(1)}, n^{(2)} , n^{(3)} | n^{(4)})
=
    R_{\mathrm{DP}}(n^{(0)}| n^{(1)}
    ) \cdot \notag \\
& \quad \left\{
    R_{\mathrm{MS}}^{\subseteq \mathrm{MS}}(n_{\mathrm{MS}}^{(1)}| n_{\mathrm{MA}}^{(2)}
    )
    R_{\mathrm{AG}}^{\subseteq\mathrm{AG}}(
    n_{\mathrm{MA}}^{(2)}| n_{\mathrm{AG}}^{(3)}
    )
    \right. \notag\\
    & \quad \left.
    +
        R_{\mathrm{ID}}^{\subseteq\overline{\mathrm{MS}}}(n_{\overline{\mathrm{MS}}}^{(1)}| n_{\overline{\mathrm{MA}}}^{(2)})
        R_{\mathrm{ID}}^{\subseteq\overline{\mathrm{AG}}}(n_{\overline{\mathrm{MA}}}^{(2)}| n_{\overline{\mathrm{AG}}}^{(3)})
 \right\}
    R_{\mathrm{UD}}(
    n^{(3)}| n^{(4)}
    ),
\label{eq:MUDecomposition}    
\end{align}
where $n^{(\cdot)}$ specifies a neuron in each local layer as
\begin{align}
n^{(0)} &\in \mcS(\{\bfeta_{v_{\mathrm{o}}^l}^{(l-1)},\bfeta_{v_{\mathrm{d}}^l}^{(l-1)}, \bfepsilon^{(e_l)}(t^-_{v_{\mathrm{o}}^l}), \bfepsilon^{(e_l)}(t^-_{v_{\mathrm{d}}^l})\}),
\notag \\
n^{(1)} &\in \text{MS} \cup \overline{\text{MS}}, n^{(1)}_{\text{MS}} \in \text{MS}, n^{(1)}_{\overline{\text{MS}}} \in  \overline{\text{MS}},
\notag \\
\text{MS} &= \mcS(\{\bfeta_{v_{\mathrm{o}}^l}^{(l-1)},\bfeta_{v_{\mathrm{d}}^l}^{(l-1)}, \bfepsilon^{(e_l)}(t^-_{v_{\mathrm{o}}^l}), \bfepsilon^{(e_l)}(t^-_{v_{\mathrm{d}}^l})\}),
\notag \\
\overline{\text{MS}} &= \mcS(\{\bfeta_{v_{\mathrm{o}}^l}^{(l-1)},\bfeta_{v_{\mathrm{d}}^l}^{(l-1)}\}),
\notag \\
n^{(2)} &\in \text{MA} \cup \overline{\text{MA}}, n^{(2)}_{\text{MA}} \in \text{MA}, n^{(2)}_{\overline{\text{MA}}} \in  \overline{\text{MA}},
\notag \\
\text{MA} &= \mcS(\{\bfm_{v_{\mathrm{o}}^l}^{e_l},\bfm_{v_{\mathrm{d}}^l}^{e_l}\}),
\label{eq:MUDecompositionNeurons} \\
\overline{\text{MA}} &= \mcS(\{\bfeta_{v_{\mathrm{o}}^l}^{(l-1)},\bfeta_{v_{\mathrm{d}}^l}^{(l-1)}\}),
\notag \\
n^{(3)} &\in \text{AG} \cup \overline{\text{AG}}, n^{(3)}_{\text{AG}} \in \text{AG}, n^{(3)}_{\overline{\text{AG}}} \in  \overline{\text{AG}},
\notag \\
\text{AG} &= \mcS(\{\widehat{\bfm_{v_{\mathrm{o}}^l}},\widehat{\bfm_{v_{\mathrm{d}}^l}}),
\notag \\
\overline{\text{AG}} &= \mcS(\{\bfeta_{v_{\mathrm{o}}^l}^{(l-1)},\bfeta_{v_{\mathrm{d}}^l}^{(l-1)}\}),
\notag \\
n^{(4)} &\in \mcS(\{\bfeta_{v_{\mathrm{o}}^l}^{(l)},\bfeta_{v_{\mathrm{d}}^l}^{(l)}\}). \notag 
\end{align}
Here, 
$R_{\mathrm{DP}}(\cdots)$ is the relevance for the DP layer duplicating $\{\bfeta_{v_{\mathrm{o}}^l}^{(l-1)},\bfeta_{v_{\mathrm{d}}^l}^{(l-1)}\}$,
$
    R_{\mathrm{MS}}^{\subseteq\mathrm{MS}}(\cdots)$
    and 
$    R_{\mathrm{AG}}^{\subseteq\mathrm{AG}}(\cdots)$
are the local relevances for the
  MeSsage (MS) and AGgregate (AG)  
   functions, respectively,
   $R_{\mathrm{ID}}^{\subseteq\overline{\mathrm{MS}}}(\cdots)$ and $R_{\mathrm{ID}}^{\subseteq\overline{\mathrm{AG}}}(\cdots)$ are the local relevance for the identical copying process for $\{\bfeta_{v_{\mathrm{o}}^l}^{(l-1)},\bfeta_{v_{\mathrm{d}}^l}^{(l-1)}\}$, and  
    $R_{\mathrm{UD}}(\cdots)$
is the relevance for the UpDate (UD) function. 

\paragraph{Message (MS) Module and Aggregate (AG) Module}

For the simplified ETGNN, the message function is identity.
The aggregate function is also identity as each batch only contains a single event. Therefore, 
\begin{align}
    &R_{\mathrm{MS}}(n^{(0)}|n^{(1)}) = R_{\mathrm{ID}}(n^{(0)}|n^{(1)}),
    &n^{(0)} &\in \mathrm{MS}, n^{(1)}\in \mathrm{MA},\\
    &R_{\mathrm{AG}}(n^{(0)}|n^{(1)}) = R_{\mathrm{ID}}(n^{(0)}|n^{(1)}),
    &n^{(0)} &\in \mathrm{MA}, n^{(1)}\in \mathrm{AG}.
\end{align}

\paragraph{Update (UD) Module}

The UD module is a GRU cell, and therefore, 
\begin{align}
    R_{\mathrm{UD}}(n^{(0)}|n^{(1)}) &= R_{\mathrm{GRU}}(n^{(0)}|n^{(1)}),\\
    n^{(0)} &\in \mcS(\{\widehat{\bfm}_{v_{\mathrm{o}}^l},\widehat{\bfm}_{v_{\mathrm{d}}^l}\bfeta_{v_{\mathrm{o}}^l}^{(l-1)},\bfeta_{v_{\mathrm{d}}^l}^{(l-1)}\}),\\
    n^{(1)} &\in \mcS(\{\bfeta_{v_{\mathrm{o}}^l}^{(l)},\bfeta_{v_{\mathrm{d}}^l}^{(l)}\}).
\end{align}

\subsection{Embedding (Emb) Module}

In the simplified ETGNN, the Emb module is an ID module that uses the final node memory as node embedding, i.e.,
\begin{align}
    \bfzeta_{v} = \bfeta_v^{(|\mcE|)}.
\end{align}
Therefore, $R_{\mathrm{Emb}}(n^{(0)}|n^{(1)}) = R_{\mathrm{PID}}(n^{(0)}|n^{(1)})$, where $n^{(0)}\in\mcS(\{\bfeta_v^{(|\mcE|)}\}_{v\in\mcV}\cup\{\bfepsilon^{(e)}(\cdot)\}_{e\in\mcE})$ and $n^{(1)}\in\mcS(\{\bfzeta_v\}_{v\in\mcV})$. $R_{\mathrm{PID}}(\cdots)$ is the relevance of the Partial IDentity (PID) function that copies a part of the input (node memory) and drops the rest part (event embedding).

\subsection{Decoding (Dec) Module}

The Dec module is a node-level MLP that transforms node embeddings to node-level predictions, i.e.,
$
    y_v = \mathrm{MLP}(\zeta_v).
$
Therefore, its relevance is $R_{\mathrm{Dec}}(n^{(0)}|n^{(1)})=R_{\mathrm{MLP}}(n^{(0)}|n^{(1)})$, $n^{(0)}\in\mcS(\{\bfzeta_v\})$, $n^{(1)}\in\mcS(\{y_v\})$.

\subsection{Relevance Definition of Basic Modules}
The relevance of the DuPlication module
$R_{\mathrm{DP}}(\cdots)$,
the IDentical copying module
$R_{\mathrm{ID}}^{ \subseteq \overline{\mathrm{MU}}} (\cdots)$,
the GRU cell module $R_{\mathrm{GRU}}(\cdots)$,
and
the Partial IDentity (PID) function module $R_{\mathrm{PID}}(\cdots)$,
can be defined analogously to that of the LSTM cell module, of which the construction is detailed in \ref{app:lstm}.

\section{Quantitative Criteria for Evaluating Explanations}
\label{sec:quant_eval_method}

We use ground-truth-based and perturbation-based metrics. For the synthetic datasets, where the ground-truth of relevant events are known, we can
compute precision and recall curves for the top-$k$ most relevant events  $\mathcal E^{\text{top}}_k$ detected by XAI methods as
\begin{align}
    \text{Precision}_k = \textstyle \frac{|\mathcal E^{\text{ground-truth}} \cap \mathcal E^{\text{top}}_k|}{k},
    \label{eq:DefPrecision}\\
    \text{Recall}_k = \textstyle  \frac{|\mathcal E^{\text{ground-truth}} \cap \mathcal E^{\text{top}}_k|}{|\mathcal E^{\text{ground-truth}}|}
    \label{eq:DefRecall},
\end{align}
where $\mathcal E^{\text{ground-truth}}$ is the set of ground-truth relevant events.
In the analysis of the infection network in \Cref{sec:infection_quantitative}, it is also important to identify possible chains of events that connect an initial infected person to the target person.  To assess the performance of such infection chain identification, we also define 
 \begin{align}
    \text{Recall-chain}_k = 
    \mathbbm{1}\left( \exists \mathcal C \in \mathfrak G \text{ s.t. } \mathcal C \subseteq {\mathcal E}_k^{\text{top}} \right),
    \label{eq:recall-chain}
\end{align}
where
 $\mathfrak{G} = \{ \mathcal C_1, \dots, \mathcal C_M \in \mathcal G \}$
 denotes the set of ground-truth 
 infection chains.
This criterion gives 1 if at least one entire ground-truth chain is included in the top-$k$ relevant events.

For any dataset,  we can use perturbation-based metrics.
We use pruning and activation curves \citep{DBLP:conf/icml/XiongSMMN22}, which measure the decrease in output probability when removing the top-$k$ relevant events from the original temporal graph, and the output probability when only using the top-$k$ relevant events as input, respectively:
\begin{align}
    \text{Prune}_k &= f^{\text{model}}_c(\mathcal E) - f^{\text{model}}_c(\mathcal E \backslash \mathcal E^{\text{top}}_k),
    \label{eq:DefPerturb}\\
    \text{Activate}_k &=  f^{\text{model}}_c(\mathcal E^{\text{top}}_k)
    \label{eq:DefActivate},
\end{align}
where $f^{\text{model}}_c(\cdot)$ is the model output probability for the target class $c$. If the events $\mathcal E^{\text{top}}_k$ are relevant, the model prediction for the target class should drop drastically when we remove these events from the input, giving a high $ \text{Prune}_k$. Similarly, predictions with input events consisting only of $\mathcal E^{\text{top}}_k$ should increase rapidly when we add the most relevant events, giving a high $\text{Activate}_k$.
Therefore, for both criteria, higher values  indicate better explanation quality.

To give an overall performance score in \Cref{tab:experiment_results}, we use the average score over the top-$k$ events.

\section{Model Training Details}
\label{sec:model_info}

For all datasets, the ETGNN has identity message function, mean aggregate function, a GRU cell for the update function, and a linear layer for the decoder.

For the Infection dataset, the dimension of node memory is set to 10, while the model output dimension is 2  for the binary classification task. 
We tested two models with the embedding function being identity (ETGNN-id) or a 1-layer attention mechanism (ETGNN-attn). Test Accuracy for ETGNN-id is 80.7\% and for ETGNN-attn is 81.6\%.

For the Attacker dataset, we use the identity embedder function,
and set the dimension of node memory again to 10.
The model's output dimension is 2. Test Accuracy is 99.0\%.

For the ICEWS18 dataset, we use the identity embedding function,
and set the dimension of node memory to 100.  The output dimension is 256. Test Accuracy is 32.7\%.

\section{Additional Emprical Results}
\label{app:qualitative_experiment}

\begin{figure*}[t]
    \centering
    \begin{subfigure}{0.91\textwidth}
        \centering
        \includegraphics[width=\linewidth]{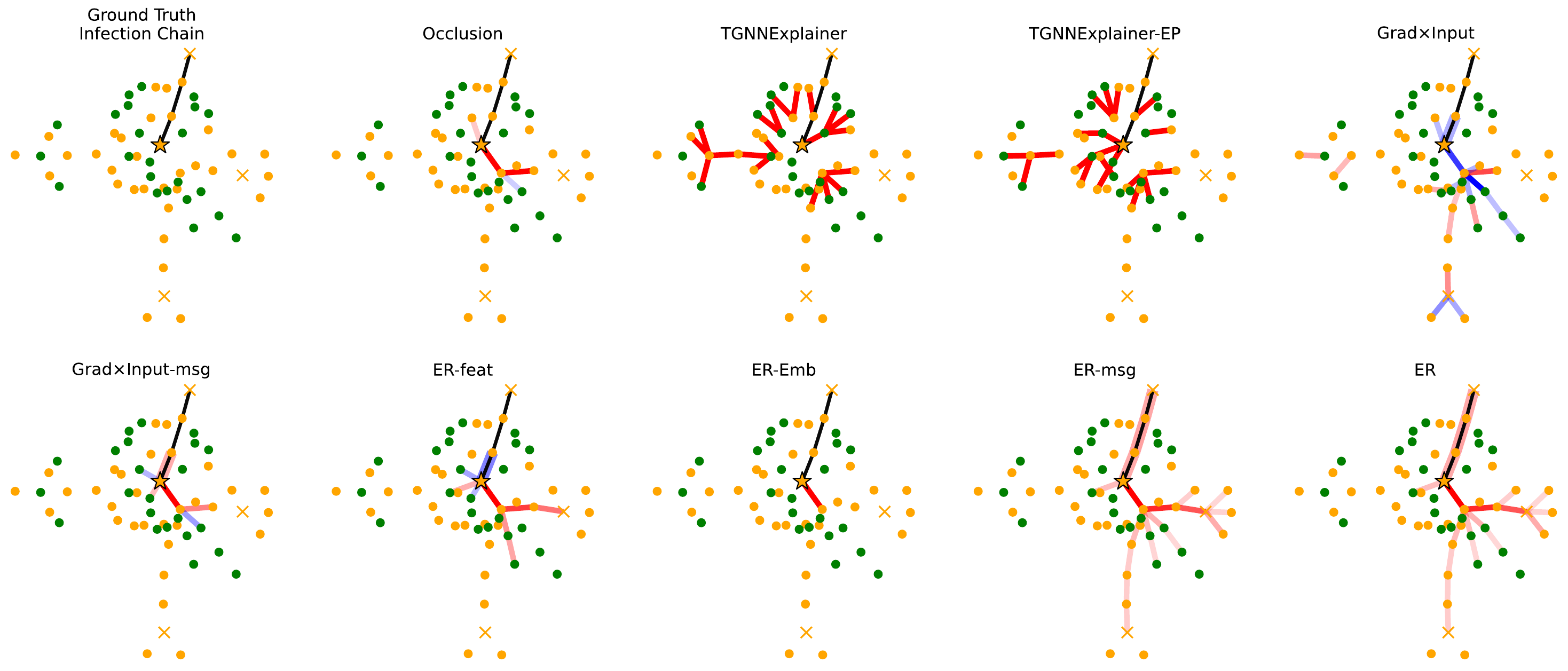}
        \caption{Prediction of Node 10.}
    \end{subfigure}

    \begin{subfigure}{0.91\textwidth}
        \centering
        \includegraphics[width=\linewidth]{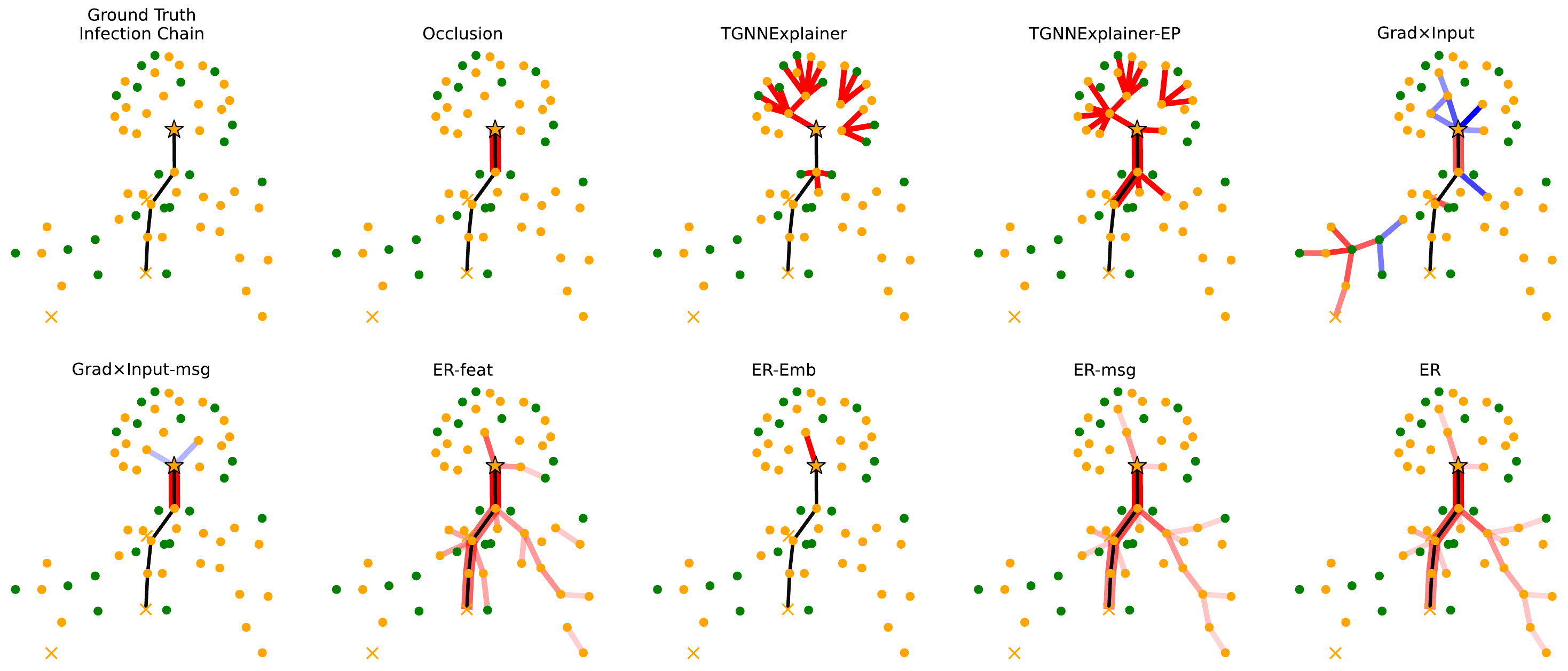}
        \caption{Prediction of Node 17.}
    \end{subfigure}
    \caption{The top-20 most relevant events for node prediction in the Infection dataset. The ground-truth infection chain depicted as black lines is the most probable one. 
The yellow crosses \textcolor{orange}{\textbf{$\times$}} 
are the initial infected nodes, and 
    the yellow star \textcolor{orange}{$\bigstar$} is the predicted node. The other yellow/green nodes are infected/not infected nodes at the end of the episode.
Overall, ER-feat, ER and ER-msg detect the infection chain, while the other methods fail.}
    \label{fig:A.extra_vis_inf}
\end{figure*}

\begin{figure}[t]
    \centering
    \includegraphics[width=\linewidth]{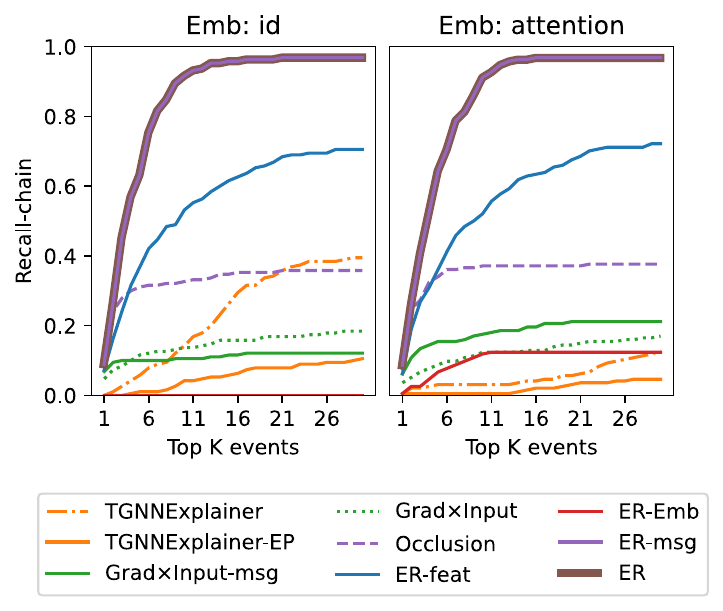}
    \caption{ 
    The Recall-chain \eqref{eq:recall-chain} with the top-$k$  relevant events for the model with the identity Emb module (left) and with the attention Emb module (right).
        ER-msg and ER perform almost identically. 
        Other methods perform poorly even for the attention Emb model, highlighting the importance of accounting for the information flow in the EP module.}
    \label{fig:infection_find_infection_chain}
\end{figure}

\begin{figure}[t]
    \centering
    \includegraphics[width=\linewidth]{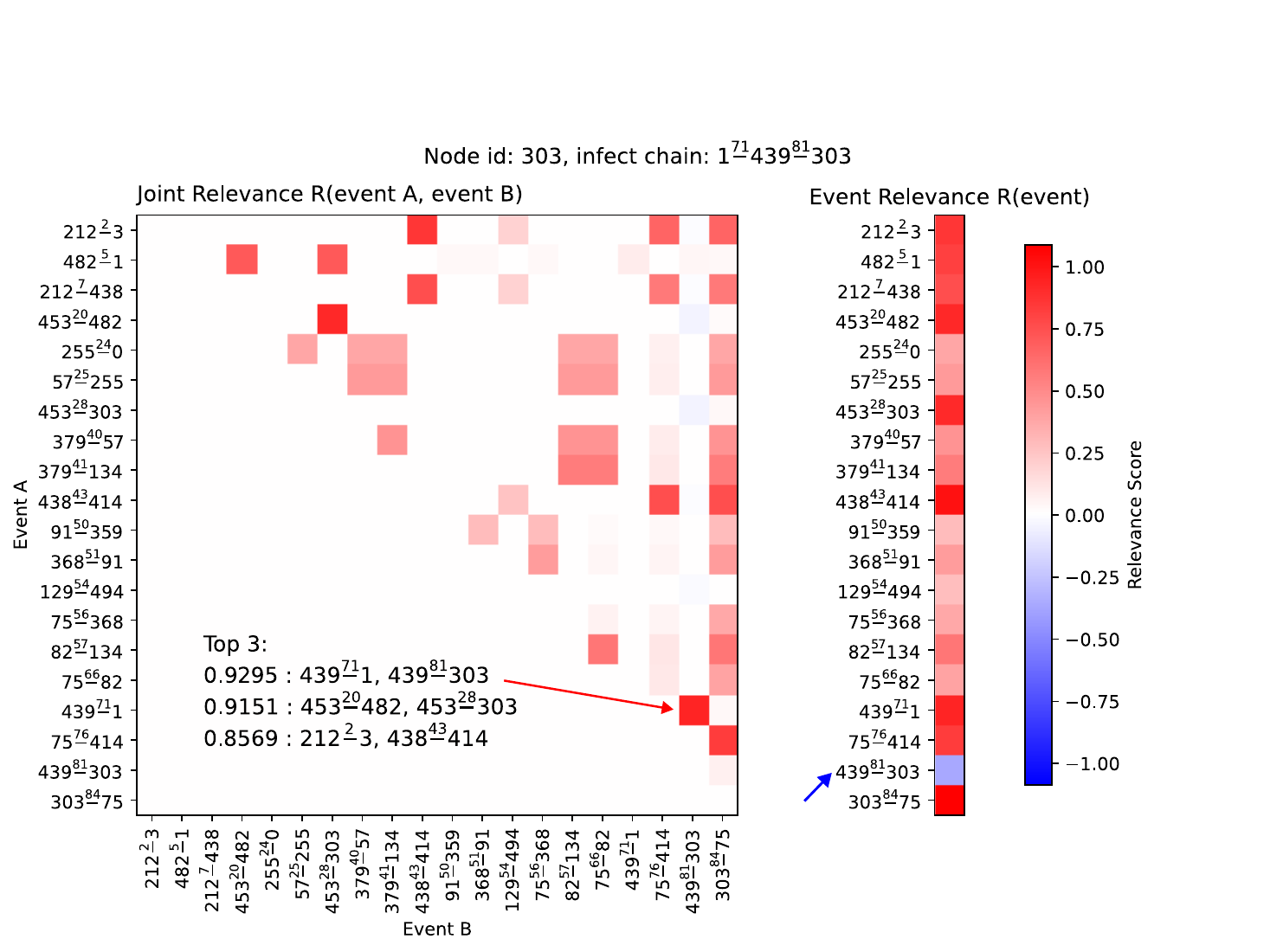}
    \caption{Joint and marginal event relevance for predicting node 303. 
    The joint relevance identified the ground-truth chain consisting of two events (439\eventtime{71}1 and 439\eventtime{81}303) as the most relevant two-event chain, while the marginal relevance assigns a negative relevance to one (439\eventtime{81}303) of those events. 
    This plot only shows 20 events that have highest absolute value of marginal relevance for the sake of clarity.}
    \label{fig:joint_rel_infection_303}
\end{figure}

\subsection{Infection Dataset}

\Cref{fig:A.extra_vis_inf} shows the top-20 most relevant events for the predictions of two additinoal target nodes not considered  in the main text. 

\Cref{fig:infection_find_infection_chain} shows the Recall-chains for ETGNNs with the identity Emb module (left) and with an attention Emb module (right).  For both models, our proposed ER and ER-msg clearly outperform the others.

\Cref{fig:joint_rel_infection_303} shows the joint relevance with respect to the prediction of another target node, where the ground-truth is an infection chain consisting of two events.  The two-event joint relevance successfully  identifies the ground-truth chain as the most relevant interaction, while single-event marginal event relevances fail to identify both events as positively relevance events---one of the links is assigned negative relevance.

\subsection{Attacker Dataset}
\label{app:attacker_experiment}

\begin{figure*}[t]
    \centering

    \begin{subfigure}{0.75\linewidth}
        \includegraphics[width=\linewidth]{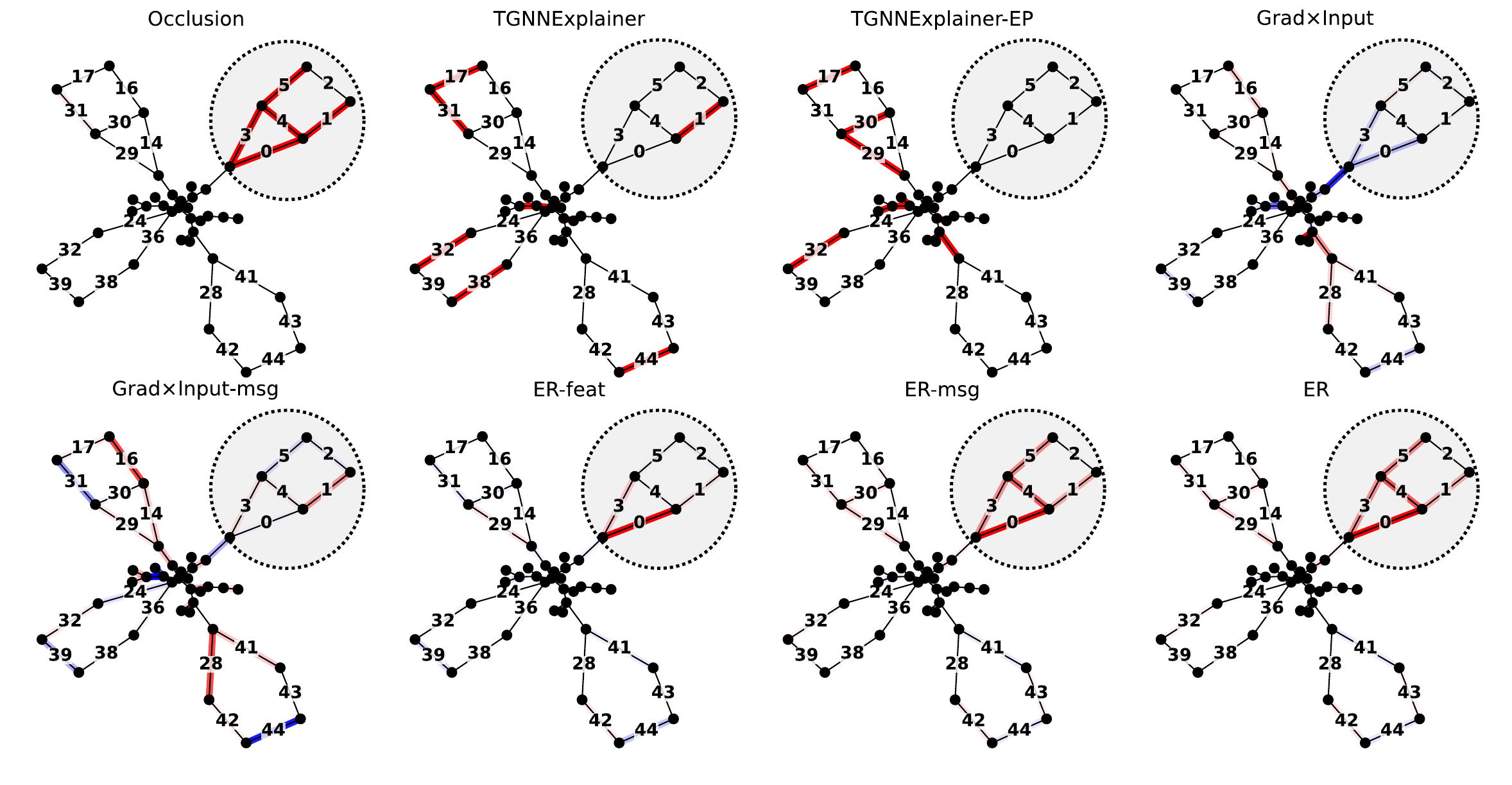}
        \vskip -0.5em
        \caption{Sample with a single attacker subgraph.
        }
        \label{fig:attacker_vis_1attacker}
    \end{subfigure}

    \begin{subfigure}{0.8\linewidth}
        \includegraphics[width=\linewidth]{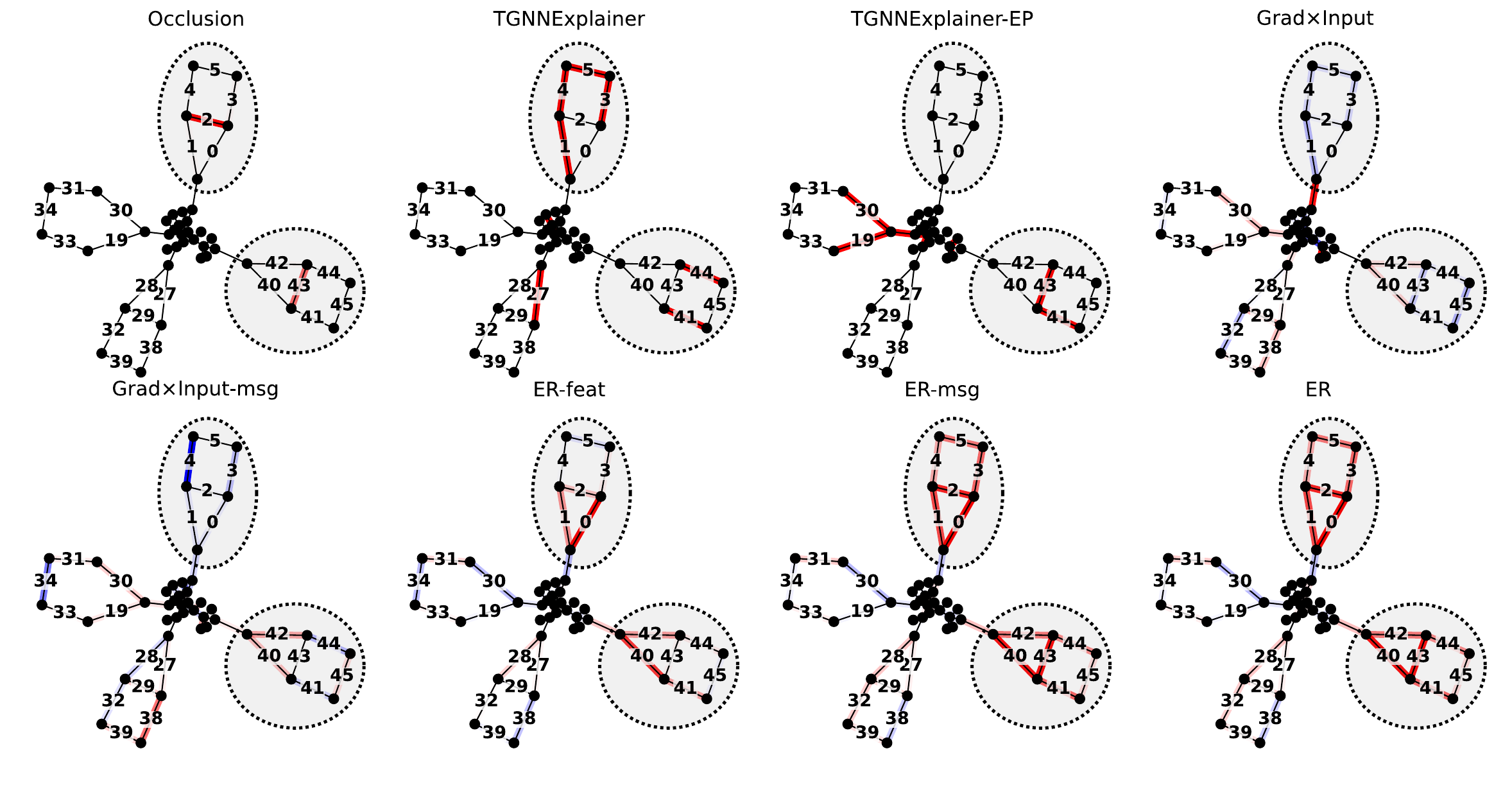}
         \caption{Sample with two attacker subgraphs. 
         }
        \label{fig:attacker_vis}
    \end{subfigure}
    
    \caption{Relevance heatmaps for a  sample with a single attacker subgraph (top) and for a sample with two attacker subgraphs (bottom).
    The attacker subgraphs are marked with circles,
    and the number on each edge denotes its timestamp.
    }
    \label{fig:app_attacker}
\end{figure*}

\Cref{fig:app_attacker} shows the relevance heatmaps obtained by 
the baseline methods (upper-row) and our proposed methods (bottom-row)
for positive samples (a) with a single attacker subgraph and (b) with two attacker subgraphs.
 Since  TGNNExplainer, as well as TGNNExplainer-EP, does not output edge-level explanation, we plot the most relevant 10 events subgraph in red with uniform intensity.
 The attacker subgraphs are marked with circles, and 
 the numbers on edges indicates the time-stamp.

We observe that 
our ER and ER-msg correctly identify the attacker subgraph as positively relevant, while the other methods fail.
Although Occlusion also explains the attacker subgraph well for the single attacker sample, as seen in \Cref{fig:app_attacker} (top), it fails to find attacker subgraphs when the sample has multiple attackers,
as seen in \Cref{fig:app_attacker} (bottom).
This behavior is as expected, because removing just one attacker should not change the prediction. 
In contrast, 
our proposed ER and ER-msg, both of which account for information flow in the EP module, successfully identify the attacker subgraphs even when multiple attackers exist.

\begin{figure*}[t]
    \centering
        \begin{subfigure}{\linewidth}
        \centering
        \includegraphics[width=0.9\linewidth]{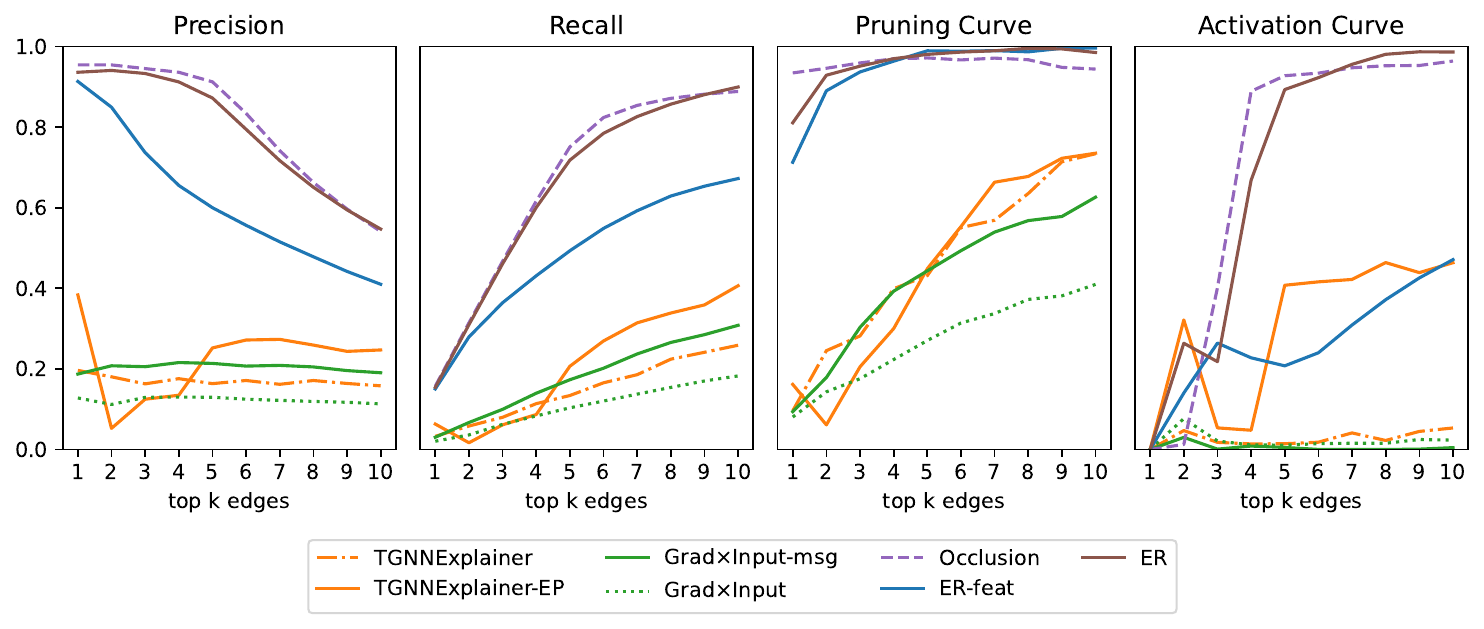}
        \vskip -0.6em    \label{fig:attacker_res_one_motif}
        \caption{On 219 correctly classified positive samples with a \textbf{single} attacker motif.}
    \end{subfigure}
    \begin{subfigure}{\linewidth}
        \centering
        \includegraphics[width=0.9\linewidth, trim=0 43 0 0, clip]{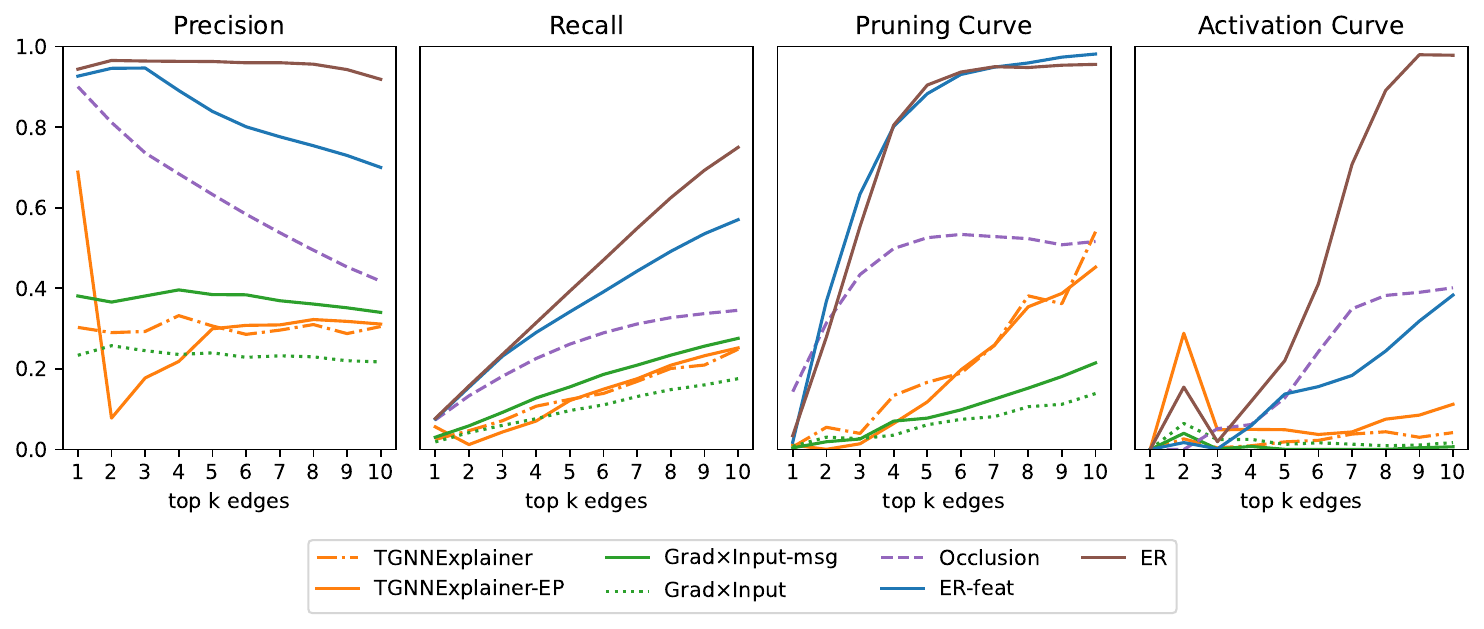}
        \vskip -0.7em        \label{fig:attacker_quantitative_multiple_motif}
          \caption{On 231 correctly classified  positive samples with \textbf{multiple} attacker motifs.}
    \end{subfigure}
    \caption{Precision, Recall, Pruning, and Activate curves on Attacker dataset.
    ER and Occlusion perform similarly good  for the single attacker cases (top), while Occlusion performs poorly  for the multiple attacker cases (bottom).}
    \label{fig:app_attacker_res_multiple_motif_vs_one_motif}
\end{figure*}

We quantitatively compare the quality of the top-$k$ most relevant events detected by our methods and the baselines.  For TGNNExplainer that attributes relevance to subgraphs, the top-$k$ relevant edges are defined as the edges of the most relevant subgraph consisting of $k$ edges.
\Cref{fig:app_attacker_res_multiple_motif_vs_one_motif} shows in each column the Precision \eqref{eq:DefPrecision}, Recall \eqref{eq:DefRecall}, Pruning \eqref{eq:DefPerturb} and Activate \eqref{eq:DefActivate} curves evaluated on the positive samples.  
We separately evaluated the performance on the set of graphs that contain one attacker subgraph (top row) and on the set of graphs that contain more than one attackers (bottom row).
For this evaluation, we augmented the training data with 
200 samples with two or more attacker motifs, so that the number of samples with a single attacker and with multiple attackers is comparable.
The results are consistent with the qualitative results 
above:
ER and Occlusion perform similarly and outperform the other methods for the single attacker samples, while Occlusion fails to identify attacker motifs for the multiple attacker samples.
These results empirically prove the superior performance of our ER methods over the baselines.

\subsection{Political Events Dataset}
\label{app:icews_exp}

We show an additional example of explaining the prediction on the event ``Shinzo Abe$-$Express intent to meet or negotiate$\rightarrow$North Korea'' on 2018-06-05. 
\Cref{tab:icews_vis_extra2} shows the 10 most relevant and the 3 least relevant events detected by ER, Grad$\times$Input, and TGNNExplainer.
ER successfully detects events between Japan, North Korea, China and US, which explain the tense situation on the Korean Peninsula at that time. 
More specifically, it finds relatively old events ($\sim 150$ days ago), e.g.,  ``North Korea$-$make statement$\rightarrow$United States'' and vise versa, that explain the political background,
as well as shortly-before events ($\sim 3$ days ago), e.g., ``Shinzo Abe$-$discuss by telephone$\rightarrow$Donald Trump'' and vise versa, that might directly triggered Shinzo Abe's expression of intent to meet or negotiate with North Korea. 
In contrast, the baseline methods yield less interpretable explanations:
Grad$\times$Input tends to find 
some irrelevant events between Donald Trump and other countries such as Australia,
while 
TGNNExplainer 
tends to focus on irrelevant events that happened one day ago, ignoring  long-range event dependencies.

\begin{table*}[t]
    \tiny
    \centering
    \begin{tabular}{R{.2\textwidth}cL{.33\textwidth}cC{.18\textwidth}|l}
    \toprule
\textbf{Method:  ER}\\
\hline
    Event &&&&& \#D ago\\
    \hline
\rowcolor{Salmon!100} \textbf{North Korea} & \textbf{--} & \textbf{Make statement} & \textbf{$\rightarrow$} & \textbf{United States} & \textbf{156} \\
\rowcolor{Salmon!98} Government (North Korea) &--& Impose administrative sanctions &$\rightarrow$& North Korea  & 156\\
\rowcolor{Salmon!97} \textbf{Donald Trump} & \textbf{--} & \textbf{Make statement} & \textbf{$\rightarrow$} & \textbf{North Korea} & \textbf{156} \\
\rowcolor{Salmon!96} Kim Jong-Un &--& Make statement &$\rightarrow$& North Korea  & 156\\
\rowcolor{Salmon!93} \textbf{Shinzo Abe} & \textbf{--} & \textbf{Discuss by telephone} & \textbf{$\rightarrow$} & \textbf{Donald Trump} & \textbf{3} \\
\rowcolor{Salmon!93} \textbf{Donald Trump} & \textbf{--} & \textbf{Discuss by telephone} & \textbf{$\rightarrow$} & \textbf{Shinzo Abe} & \textbf{3} \\
\rowcolor{Salmon!87} \textbf{China} & \textbf{--} & \textbf{Host a visit} & \textbf{$\rightarrow$} & \textbf{Shinzo Abe} & \textbf{154} \\
\rowcolor{Salmon!86} \textbf{Shinzo Abe} & \textbf{--} & \textbf{Make a visit} & \textbf{$\rightarrow$} & \textbf{China} & \textbf{154} \\
\rowcolor{Salmon!66} Shinzo Abe &--& Make statement &$\rightarrow$& Government (Japan)  & 5\\
\rowcolor{Salmon!62} North Korea &--& Express intent to meet or negotiate &$\rightarrow$& South Korea  & 156\\
$\cdots$\\
\rowcolor{cyan!1} Donald Trump &--& Consult &$\rightarrow$& Sheikh Hamad bin Isa al Khalifah  & 156\\
\rowcolor{cyan!2} Shinzo Abe &--& Engage in diplomatic cooperation &$\rightarrow$& Donald Trump  & 27\\
\rowcolor{cyan!2} Donald Trump &--& Engage in diplomatic cooperation &$\rightarrow$& Shinzo Abe  & 27\\
    \bottomrule
\\
\textbf{Method:  Grad$\times$Input}\\
\hline
    Event &&&&& \#D ago\\
    \hline
\rowcolor{Salmon!100} Donald Trump &--& Make statement &$\rightarrow$& Australia  & 123\\
\rowcolor{Salmon!99} Ministry (Iran) &--& Make statement &$\rightarrow$& Donald Trump  & 123\\
\rowcolor{Salmon!80} Donald Trump &--& Express intent to engage in diplomatic cooperation (such as policy support) &$\rightarrow$& Malcolm Bligh Turnbull  & 123\\
\rowcolor{Salmon!46} Donald Trump &--& Make statement &$\rightarrow$& North Korea  & 156\\
\rowcolor{Salmon!43} Donald Trump &--& Threaten to reduce or stop aid &$\rightarrow$& Pakistan  & 156\\
\rowcolor{Salmon!32} Donald Trump &--& Make statement &$\rightarrow$& Bharatiya Janata  & 156\\
\rowcolor{Salmon!25} Shinzo Abe &--& Make statement &$\rightarrow$& Indonesia  & 136\\
\rowcolor{Salmon!24} North Korea &--& Express intent to meet or negotiate &$\rightarrow$& South Korea  & 156\\
\rowcolor{Salmon!22} Kim Jong-Un &--& Make statement &$\rightarrow$& North Korea  & 156\\
\rowcolor{Salmon!22} China &--& Mobilize or increase police power &$\rightarrow$& Vietnam  & 156\\
$\cdots$\\
\rowcolor{cyan!24} Donald Trump &--& Make pessimistic comment &$\rightarrow$& Pakistan  & 156\\
\rowcolor{cyan!96} China &--& Host a visit &$\rightarrow$& Shinzo Abe  & 154\\
\rowcolor{cyan!100} Shinzo Abe &--& Make a visit &$\rightarrow$& China  & 154\\
    \bottomrule
\\
\textbf{Method:  GNNExplainer}\\
\hline
    Event &&&&& \#D ago\\
    \hline
\rowcolor{Salmon!50} Shinzo Abe &--& Make statement &$\rightarrow$& Aso Taro  & 1\\
\rowcolor{Salmon!50} Vladimir Putin &--& Express intent to meet or negotiate &$\rightarrow$& North Korea  & 1\\
\rowcolor{Salmon!50} South Korea &--& Make an appeal or request &$\rightarrow$& North Korea  & 1\\
\rowcolor{Salmon!50} United States &--& Make an appeal or request &$\rightarrow$& North Korea  & 1\\
\rowcolor{Salmon!50} Shinzo Abe &--& Meet at a 'third' location &$\rightarrow$& G7  & 1\\
\rowcolor{Salmon!50} G7 &--& Meet at a 'third' location &$\rightarrow$& Shinzo Abe  & 1\\
\rowcolor{Salmon!50} Shinzo Abe &--& Make an appeal or request &$\rightarrow$& Ministry (Japan)  & 1\\
\rowcolor{Salmon!50} North Korea &--& Return, release person(s) &$\rightarrow$& Citizen (United States)  & 1\\
\rowcolor{Salmon!50} North Korea &--& Host a visit &$\rightarrow$& Sergey Viktorovich Lavrov  & 1\\
\rowcolor{Salmon!50} Sergey Viktorovich Lavrov &--& Make a visit &$\rightarrow$& North Korea  & 1\\
    \bottomrule
    \end{tabular}
    \vspace{.3cm}
    \caption{The most relevant events by ER (top), Grad$\times$Input (middle) and TGNNExplainer (bottom). `$\#D$ ago' indicates how many days the event happened before the predicted event: Shinzo Abe$-$Express intent to meet or negotiate$\rightarrow$North Korea@2018-06-05.}
    \label{tab:icews_vis_extra2}
\end{table*}

\begin{figure}[t]
    \centering
    \includegraphics[width=0.9\linewidth]{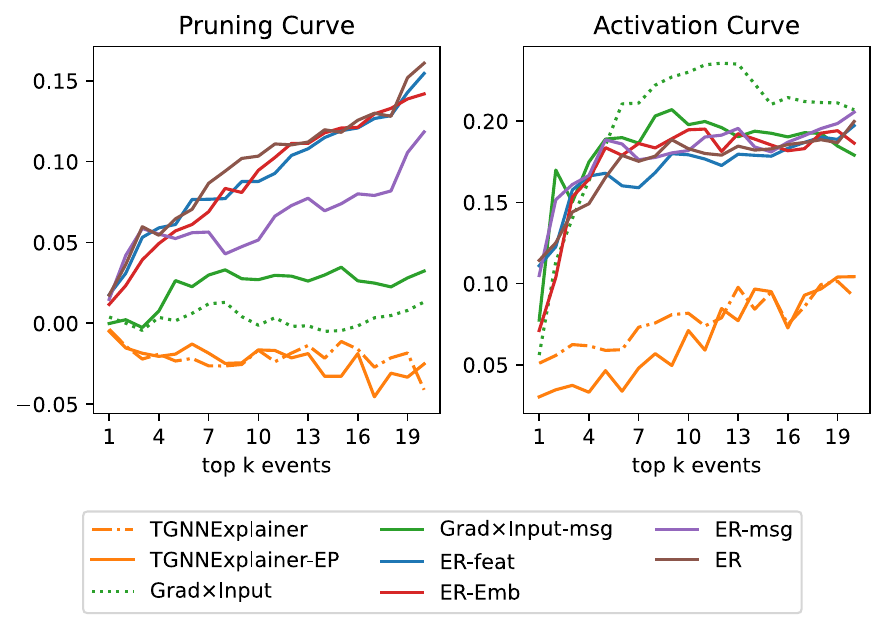}
    \caption{
    Pruning and Activation curves on ICEWS18. 
    }
    \label{fig:pruning_eval_icews}
\end{figure}

\Cref{fig:pruning_eval_icews} shows the results of pruning and activation tests for the top-20 most relevant events on 50 randomly sampled events that are correctly predicted.
We observe from the pruning curves that our ER, ER-feat, and ER-Emb outperform the others. 
The activation curves show a similar trend except 
for Grad$\times$Input, which performs well for large $k$.
Overall, 
our ER methods outperform the baseline methods.

\end{document}